\documentclass{article}

\usepackage[preprint]{neurips_2025}

\usepackage[T1]{fontenc}
\usepackage[utf8]{inputenc}

\usepackage{amsmath,amssymb,mathtools,bbm}

\usepackage{graphicx}
\usepackage{booktabs,multirow,tabularx,threeparttable}
\usepackage{subcaption}   
\usepackage{siunitx}
\usepackage{algorithm}
\usepackage{algpseudocode}

\usepackage{float}
\usepackage{adjustbox}
\usepackage{placeins}
\usepackage{enumitem}
\usepackage{tikz}
\usetikzlibrary{arrows.meta,positioning,shapes.geometric,calc}
\usepackage[hyphens]{url}  

\usepackage{listings}

\lstdefinestyle{prompt}{
  basicstyle=\ttfamily\small,
  breaklines=true,
  breakatwhitespace=false,   
  columns=fullflexible, keepspaces=true,
  frame=single, framesep=6pt,
  inputencoding=utf8,
  literate=
   {–}{{\textendash}}1
   {—}{{\textemdash}}1
   {“}{{``}}1
   {”}{{''}}1
   {’}{{'}}1
   {…}{{\dots}}1
}
\newcolumntype{Y}{>{\centering\arraybackslash}X}

\usepackage{microtype}
\usepackage{nicefrac}
\usepackage{xcolor}

\usepackage{hyperref}

\title{Do Large Language Models (LLMs) Understand Chronology?}

\author{%
  Pattaraphon Kenny Wongchamcharoen\\
  University of California, Berkeley\\
  Berkeley, CA 94720\\
  \texttt{pattaraphon.kenny@berkeley.edu}
  \And                                
  Paul Glasserman\\
  Columbia Business School\\
  New York, NY 10027\\
  \texttt{pg20@gsb.columbia.edu}
  }

\begin{document}

\maketitle

\begin{abstract}
Large language models (LLMs) are increasingly used in finance and economics, where prompt-based attempts against look-ahead bias implicitly assume that models understand chronology. We test this fundamental question with a series of chronological ordering tasks with increasing complexities over facts the model already knows from pre-training. Our tasks cover (1) chronological ordering, (2) conditional sorting (filter, then order), and (3) anachronism detection. We evaluate GPT-4.1 and Claude-3.7 Sonnet, with and without Extended Thinking (ET), and the newly released GPT-5 across multiple reasoning-effort settings. Across models, Exact match rate drops sharply as sequences lengthen even while rank correlations stay high as LLMs largely preserve local order but struggle to maintain a single globally consistent timeline. In conditional sorting, most failures stem from the filtering step rather than the ordering step, but GPT-5 and Claude-3.7 with Extended Thinking outshine normal models significantly. Lastly, anachronism detection is found to be the easiest task for the LLMs but performance still declines with increasingly overlapping timelines or entities. Overall, our main significant contribution is showing that allocating explicit reasoning budget helps with chronological ordering with GPT-5 at medium/high reasoning effort achieving flawless ordering at all lengths and perfect conditional sorting (both self-filtered and given-subset), whereas low/minimal effort degrades with longer lists, mirroring earlier models. We release all code and evaluation templates to support full reproducibility.\footnote{This manuscript is an extended version of our work presented at the AAAI-26 AI4TS Workshop (poster) and the AAAI-26 Student Abstract Program (oral). The code repository is available at: \url{https://github.com/kennywong524/chronollm}}”
\end{abstract}

\section{Introduction}
Large language models (LLMs) have shown great potential as forecasting tools for finance and economics. Typical applications ask LLMs to predict the direction of stock prices based on news reports, company earnings calls, or analysts’ research. Yet it has also been recognized that the backtesting of LLM forecasts is subject to look-ahead bias (\citep{glasserman2024large}, \citet{SarkarVafa2024}) when the backtesting period overlaps with the LLMs training window: the LLM may be asked to predict an outcome it has already seen. Leakage of post-event information embedded in the LLM’s pre-training corpus can inflate estimated forecast performance and lead to disappointing out-of-sample results. 

Various methods have been proposed to try to measure and mitigate this problem. The simplest approach is to limit testing to an LLM’s post-training period (as in, e.g., \citet{Halawi}, \citet{LopezliraTang}), but this severely limits the testing window, particularly for the most recent models. \citet{glasserman2024large} find that masking company names in news headlines can be effective in removing look-ahead bias; this idea has been extended to larger texts in \citet{EngelbergManela2025Neutering}, but \citet{SarkarVafa2024} and \citet{LopezliraTangZhu} find evidence that LLMs can see through anonymization in large documents. Building snapshot models trained using only text available up to fixed dates in the past (\citet{SarkarVafa2024}, \citep{HeLvManelaWu2025ChronoConsistent}) provides a secure way to wall off future information, but it is computationally demanding and limited to using training documents with clear time stamps.

From a user’s perspective, the most convenient solution would be to wall off future information by instructing an LLM to respond using only information available up to a fixed date. \citet{SarkarVafa2024} and \citet{LopezliraTangZhu} find evidence of leakage in examples of this approach. More basically, a prompt-based instruction like ``use only information from before 2016'' presupposes that an LLM understands what “before 2016” means and can respond accordingly. So here we step back and ask a more fundamental question: \emph{Do LLMs understand chronology?} Before asking an LLM to avoid leakage of future information in responding to new information, we want to evaluate how well the LLM understands chronological constraints in data on which it has been trained.

Prior approaches to NLP temporal testbeds (e.g., TimeQA~\citep{timeqa}, TRAM~\citep{tram}, TimeBench~\citep{timebench}, and ChronoSense~\citep{chronosense}) seek to isolate and evaluate specific components of temporal reasoning. To this end, they evaluate an LLM’s performance on new, sometimes hypothetical scenarios designed to probe a very specific type of inference. Here we take a different approach, evaluating chronological understanding of \emph{information the LLM already knows}; we use historical facts, which we first confirm the LLM knows. This design probes the LLM’s ability to integrate temporal reasoning with the model’s internal world view. It is less focused on isolating granular components of temporal reasoning and seeks instead to better assess chronological reasoning “in the wild.”

We test performance on three task types:
\emph{(1) Chronological sorting} (putting scrambled events in chronological order);
\emph{(2) Conditional sorting} (selecting events that meet a specified condition and ordering those chronologically);
\emph{(3) Anachronism detection} (distinguishing possible vs.\ impossible events based on historical data).

We ask the LLMs to respond in a structured JSON format for us to easily evaluate their temporal abilities using different quantifiable metrics such as Spearman’s \(\rho\), Kendall’s \(\tau\) correlation coefficient, and exact match rates. Such a design requires no auxiliary user information at inference time: The model answers in a structured JSON format, allowing evaluation via rank- and set-based metrics (e.g., Kendall’s~\(\tau\), Spearman’s~\(\rho\), precision/recall/F\(_1\) for feasibility). We evaluate both non-reasoning and reasoning-mode LLMs from multiple families (e.g., OpenAI’s GPT-4.1, the newly released GPT-5 series with various reasoning effort hyperparameters and Anthropic's Claude Sonnet 3.7 with and without extended-thinking), controlling for temperature and prompting. We report aggregate accuracy for exact match rate, rank correlations for ordering, and ablations (e.g., list length, position effects, and extended-thinking) to diagnose error modes. 

Our main finding is that LLM performance on our chronology tasks degrades quickly with problem complexity. This finding does not bode well for prompt-based mitigation of look-ahead bias using current models, as the complexity of walling off future information for a forecasting task is likely to be much greater than the complexity of our experiments. However, performance improves substantially with models with extended reasoning. By contrast, the LLM performance on anachronism detection is significantly better: accuracy is high (more than 0.90) in all task sizes with near-perfect precision and recall; degradation appears only when experiments with the notion of multiple temporal windows or multi-figure overlaps are introduced.

Interestingly, poor performance at high complexity kicks in at problem scales much smaller than those seen in other tasks (\citet{Shojaee2025IllusionThinking}), suggesting that reasoning about chronology of real events may be inherently difficult. Also, complexity does not have a simple relationship with list length: for example, LLMs do better listing all U.S. presidents in chronological order than they do sorting a random list of 20 U.S. presidents.

Our results include several novel findings: (i) Exact match ordering collapses as lists grow, even while rank correlations stay fairly high; (ii) single-prompt conditional sorting fails almost completely at the filtering step, so rank correlations on valid subset can overstate performance; (iii) turning on explicit deliberation flips this pattern, yielding stable filtering and near-perfect ordering; (iv) errors often concentrate in the middle of lists, with more-salient starting and end points acting as anchors; (v) a base anachronism detection task is easy for the model, but in tests based on the intersection of multiple events, we see more false negatives as the model struggles to correctly identify the intersection of multiple time lines.

Together, today’s models show a workable but brittle sense of chronology—decent local alignment, weak global consistency—unless we force them to think using reasoning models.

\section{Methodology}

We design our tests based on widely available historical timelines, including the terms of U.S. presidents. These data sources allow the design of tasks of widely varying complexity based on information solidly within an LLM's training corpus. In particular, we use
the following:

\textbf{(a) U.S.\ presidents:} 43 individuals (excluding Grover Cleveland and Donald Trump, who served non-consecutive terms), with attributes such as birth state, first-name initial, and calendar features (birth-year parity, month, day of week).
\newline
\textbf{(b) Historical timelines:} 20th century world events drawn from curated Wikipedia timeline sources to broaden basic sorting beyond the presidents domain.

What distinguishes our design from other studies is that we test chronological reasoning over facts the model already knows. Before any ordering or detection trial, we verify item-level knowledge with a separate query: for each candidate item $i$, the model is asked to produce a \emph{year} ($\hat y_i$) — the start year of a presidency (presidents) or the event year (historical timelines). An item passes this screen if and only if $\hat y_i$ exactly matches our canonical year $y_i$. We then construct each trial list using only items that the LLM successfully filters and \emph{only then} prompt the model to order them. All results reported in the paper are computed on these knowledge-verified lists, isolating ordering skill from internal knowledge gaps. All prompt templates (for verification, basic sorting, conditional sorting, and anachronism detection) are provided in Appendix ~\ref{app:prompts}.

Our testbed comprises three task families of increasing difficulty:

\paragraph{1. Basic sorting}
Given a shuffled list, return items in strict chronological order. Evaluated on \emph{both} corpora:
\begin{itemize}[nosep,leftmargin=1em]
  \item \textbf{Presidents:} with random list of length $n\in\{2,5,10,15,20,25,30, 35,40,43\}$.
  \item \textbf{Historical timelines:} event lists of varying lengths to probe scaling and domain variety.
\end{itemize}
To test generalizability outside of the 20th century events, we also conducted a \emph{timescale} variant which samples events whose dates differ by wider time window (at least 50-100 years) to probe coarse-granularity reasoning.

\paragraph{2. Conditional sorting}
The model \emph{first} applies a Boolean filter and \emph{then} sorts the surviving presidential names. Filters use widely known attributes:
\begin{enumerate}[label=(\alph*),nosep,leftmargin=50pt]
  \item \textbf{Birth state}: Virginia, Ohio, Massachusetts, or \textsc{OhioOrVirginia};
  \item \textbf{Name prefix}: first names starting with A, B, or C;
  \item \textbf{Calendar attributes}: even birth year, day of week, month of birth, etc.
\end{enumerate}
We test two prompting paradigms: \emph{self-filter-and-sort} (single prompt, filter and order) and \emph{given-subset-sort} (subset supplied, order only).

For basic and conditional sorting, we report Exact match rate (EMR), position-wise accuracy, and three rank-based metrics—Spearman’s \(\rho\), Kendall’s \(\tau\), and normalized Cayley distance—computed after subset identification. Formal definitions appear in Appendix ~\ref{app:metrics}.

\paragraph{3. Anachronism detection}
Each trial presents a configurable number of president–event statements, and other variants such as overlapping lifetime timelines of 2-4 presidents. The model labels each statement \textsc{Possible} or \textsc{Not possible} chronologically. We record classification accuracy, precision, recall, and F-1 score, in addition to the four ranking metrics.

Our experimental design provides a concise yet flexible testbed for chronology and temporal reasoning for a couple of reasons:

\begin{enumerate}[leftmargin=1.4em,itemsep=2pt,topsep=2pt]
	\item \textbf{No external documents at inference.} Measures intrinsic temporal coherence of the pretraining-induced world model (relevant to look-ahead bias).

	\item \textbf{Controllable difficulty.} The notion of list sizes and number of events and figures, conditional filters, and number of timelines overlaps in anachronism detection provide tunable complexities and allow for position-wise error profiles analysis.

	\item \textbf{Reproducibility.} Deterministic settings (e.g., temperature = 0) with multi-seed repeats, public code, and fixed ground truth constructed from canonical timelines.
\end{enumerate}

\section{Related Work}

\subsection{LLMs and look-ahead bias in finance}
Prior work documents that LLMs may be subject to look-ahead bias in return-prediction settings, even under anti-leakage prompts \citep{glasserman2024large,SarkarVafa2024}. Mitigations include prompt engineering and masking \citep{glasserman2024large,EngelbergManela2025Neutering}, which reduce leakage but may degrade semantics or add inference cost. We refer to the Introduction for details; here we simply note that our evaluation is orthogonal: 
we want to measure chronological understanding of facts a model already knows, rather than to measure or prevent leakage of future information.

\subsection{Chronologically consistent LLMs}
Chronology-aware pretraining strategies (e.g., time-sliced vintages) mitigate leakage by construction \citep{SarkarVafa2024,HeLvManelaWu2025ChronoConsistent}, but require heavy curation and retraining for each cutoff. Our approach complements these by testing existing off-the-shelf models empirically as end users to determine whether they exhibit usable chronological coherence, whether explicit reasoning improves it, and whether an added reasoning budget helps, without any retraining required.

\subsection{Broad temporal-reasoning benchmarks}
Existing testbeds (TRAM, TimeBench, ChronoSense) primarily evaluate pairwise relations or short sequences \citep{tram,timebench,chronosense}. We avoid re-surveying them here; instead we focus on what they leave open: how accuracy scales with list length in ordering tasks with increasing complexity, and whether structured deliberation (e.g., extended thinking \citep{anthropicET}) measurably improves performance.

\section{Basic Sorting on Events} \label{sec:baseline}

We begin our experiment by establishing the foundational methodology for evaluating LLM temporal reasoning capabilities through a systematic approach to chronological ordering tasks. We start by constructing a  dataset of historical events from the 20th century, extracted from Wikipedia timeline pages using a custom parser that handles various date formats including full dates (DD/MM/YY), month and year (MM/YY), and year-only (YY) entries. This extraction process involves cleaning and validating the data through date standardization, event deduplication, and ordinal ranking assignment, resulting in a final dataset of over 1000 events with chronological order. The 20th-century timeline corpus above was then used to prototype sampling, prompts, and scoring.

The experimental design employs a systematic approach to sample size selection, testing LLMs across small sizes (2-5 events), medium sizes (10-30 events), and large sizes (35-100 events) to not only assess the model's chronological ordering ability but also to test scaling behavior. Each experiment consists of 20 trials per sample size for statistical significance, with complete random sampling without replacement and random permutation of the ground truth order. 

We use a standardized prompt structure that positions the LLM as an expert historian specializing in accurate chronological sequencing, with clear task rules requiring strict chronological ordering from earliest to latest events. The post-processing pipeline converts raw LLM response JSON files to standardized CSV format through fuzzy matching algorithms using FuzzyWuzzy string similarity with an 85\% threshold, handling missing items and extra items appropriately.

\newpage
\paragraph{Event–knowledge filtering}
Before parsing the prompts to the LLM to order our shuffled events, we first verify that it \emph{knows} each item via a separate knowledge verification query (in Appendix \ref{app:knowledgeverification}).

The filtering approach ensures we only test chronological reasoning on events GPT demonstrably knows, providing a fair evaluation of temporal reasoning capabilities. For every candidate item we issued a single query (``In what year did \emph{E} occur?'') and compared the response to the ground‑truth year.  Events for which GPT‑4.1’s answer was incorrect were removed. Among the correctly identified events, they were sampled without replacement, yielding a corpus of \(N=\,100\) items, one per calendar year of the twentieth century from the original dataset. Ordering tasks were then conducted on this filtered set so that subsequent error analyses would not confound chronological reasoning with knowledge gaps.

\tikzset{
  mybox/.style      = {rectangle, rounded corners, draw=black!60,
                       fill=gray!10, font=\footnotesize, align=center,
                       inner sep=4pt},
  mydecision/.style = {diamond, aspect=2, draw=black!60,
                       fill=gray!10, font=\footnotesize, align=center},
  myarrow/.style    = {-{Stealth[length=2mm,width=2mm]}, thick},
}
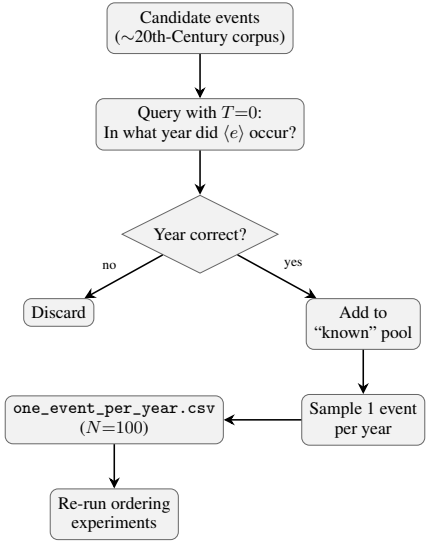
\begin{figure}[htbp]
  \centering
  \resizebox{0.4\columnwidth}{!}{%
      \begin{tikzpicture}[node distance = 8mm and 14mm]
    
        \node[mybox] (pool)
          {Candidate events\\($\sim$20th‑Century corpus)};
    
        \node[mybox, below=of pool] (prompt)
          {Query with $T{=}0$:\\\small In what year did $\langle e\rangle$ occur?};
    
        \node[mydecision, below=of prompt] (check) {Year correct?};
    
        \node[mybox, below left = 8mm and 12mm of check] (discard) {Discard};
    
        \node[mybox, below right = 8mm and 12mm of check] (keep)
          {Add to\\``known'' pool};
    
        \node[mybox, below=of keep] (sample)
          {Sample 1 event\\per year};
    
        \node[mybox, left=of sample] (csv)
          {\texttt{one\_event\_per\_year.csv}\\($N{=}100$)};
    
        \node[mybox, below=of csv] (repeat)
          {Re‑run ordering\\experiments};
    
        \draw[myarrow] (pool)   -- (prompt);
        \draw[myarrow] (prompt) -- (check);
    
        \draw[myarrow] (check)  -- node[above left, font=\scriptsize]{no}  (discard);
        \draw[myarrow] (check)  -- node[above right, font=\scriptsize]{yes} (keep);
    
        \draw[myarrow] (keep)   -- (sample);
        \draw[myarrow] (sample) -- (csv);
        \draw[myarrow] (csv)    -- (repeat);
    
      \end{tikzpicture}%
  } 
  \caption{Event‑knowledge filtering pipeline}
  \label{fig:event-filter-flow}
\end{figure}

\paragraph{Sanity check against chance}
To contextualize performance, we benchmarked GPT-4.1 against \(1{,}000\) uniformly random permutations per list length and expressed model performance as an empirical percentile relative to this distribution. Across all list sizes GPT‑4.1 ranks in the 95th–100th percentile for rank‑correlation and normalized Cayley distance, confirming a substantial advantage over random guessing. Exact match, being an “all‑or‑nothing’’ metric, converges to the 50th percentile once \(n\ge20\) because random permutations almost never achieve a perfect match. Despite this, we still see GPT-4.1's substantial edge over random guessing for Exact match when \(n\le20\) Full details and plots appear in Appendix \ref{app:rand-baseline}.

\paragraph{Wide-gap variant}
Finally, we test whether large temporal spacing between events makes ordering easier. We keep the experimental design identical but resample events so that most of them are separated by wide gaps (50–200 years). 
We curated a small pool \(\mathcal{U}\) of single-year events spanning Years \(1\)–\(2025\) (political, scientific, cultural). As in all experiments, we applied \textit{knowledge verification} first: an event \(e_k\) is retained if and only if the model returns the exact canonical year \(\hat y_k=y_k\). We then sampled lists of size \(n\in\{2,5,10,15,20,24\}\) from the gated set \(\mathcal K\), targeting wide gaps by requiring that a large share of pairs exceed a soft threshold \(\Delta_{\text{tgt}}\in\{50,100,200\}\). We keep all inference settings and evaluation metrics unchanged. Full experimental design for this variant can be found in Appendix ~\ref{app:timescale}.

\paragraph{Note on determinism and reproducibility}
All ordering experiments used a deterministic event shuffle and temperature fixed at zero (temp = 0) for our API calls. Even so, re-submitting the exact same prompt occasionally produced small differences in the ranked list—behavior commonly reported for current LLM APIs (due to their non-deterministic nature) and not specific to our setting. To accommodate this, we ran two repeats per trial and report evaluation metrics (Spearman’s~\(\rho\), Kendall’s~\(\tau\), \(\text{Cayley}_{\mathrm{norm}}\)) averaged across repeats; our qualitative results and conclusions are unchanged.

\subsection{Findings}

\paragraph{20th Century Historical Timeline Result}

We conducted 20 trials for each list size according to the methodology described in Figure \ref{fig:event-filter-flow} . Table \ref{tab:filtered-agg} shows the aggregate result for each list size, \(n\).

\begin{table}[htbp]
  \caption{Aggregated performance on the filtered corpus (20 trials per list size).}
  \label{tab:filtered-agg}
  \centering
  \begin{tabular}{
    S[table-format=3.0]
    S[table-format=1.3]
    S[table-format=1.3]
    S[table-format=1.3]
    S[table-format=2.2]
    S[table-format=1.2]
  }
    \toprule
    {$n_{\text{events}}$} & {\(\bar\rho\)} & {\(\bar\tau\)} &
    {Normalized Cayley} & {Cayley} & {Exact match rate} \\
    \midrule
      2  & 1.000 & 1.000 & 0.000 &  0.00 & 1.00 \\
      5  & 0.940 & 0.880 & 0.150 &  0.60 & 0.45 \\
     10  & 0.955 & 0.876 & 0.267 &  2.40 & 0.10 \\
     20  & 0.963 & 0.869 & 0.418 &  7.95 & 0.00 \\
     50  & 0.928 & 0.808 & 0.762 & 37.35 & 0.00 \\
    100  & 0.786 & 0.661 & 0.909 & 90.00 & 0.00 \\
    \bottomrule
  \end{tabular}
\end{table}

The most striking feature of these results are the Exact match rates in the last column. The LLM consistently orders a pair of events correctly, confirming the validity of the data and the task assignment. But with five events, the LLM achieves correct chronological ordering only about half the time. With longer lists, the LLM virtually never achieves a correct ordering. Performance on the chronological ordering task drops precipitously with problem complexity even at levels that use a fairly small number of tokens. 

Spearman's \(\rho\) and Kendall's \(\tau\) are more forgiving as they seek to quantify the quality of an imperfect sort. These measures stay relatively flat from 5 to 50, but show a marked drop at 100. Kendall’s \(\tau\) is arguably the more informative of the two measures because it is based on pairwise comparisons of the order of events.
The normalized Cayley distance rises monotonically with \(n\), capturing the increasing number of swaps required to repair longer permutations,
Taken together, these results indicate that GPT‑4.1 excels at ordering very short lists but struggles with longer lists.

We can also profile position-specific errors using the mean absolute rank difference (MARD). Figure \ref{fig:pos‑effect} plots \(\mathrm{MARD}_n(k)\) across ground-truth positions \(k\) for list lengths \(n\in\{2,5,10,20,50\}\).
The formal definition is in Appendix ~\ref{app:metrics:positionwise}; lower values indicate smaller displacement. We observe that for short lists (\(n{=}2,5,10\)), MARD remains below two positions for \emph{all} ground‑truth slots, and the shaded error bands are narrow, indicating stable accuracy regardless of where an event appears. However, for longer lists (\(n{=}20,50\), errors grow to 3–6 positions on average and vary sharply with position; variability bands widen, showing that the model’s placement becomes both less accurate and less reliable.

\paragraph{Wide-Gap Variant Result}

Figure~\ref{fig:timescale-metrics} and Table~\ref{tab:timescale-stats} summarize performance when events are separated
by centuries and the pool is pre‑filtered to facts the model knows. Rank correlations drop slightly, but remain high as list length grows: for \(n\in\{10,15,20,24\}\) we observe \(\rho\approx 0.96\!-\!0.97\) and \(\tau\approx 0.89\!-\!0.92\). However, strict exact‑match collapses beyond \(n\!\approx\!10\). Mean exact‑match falls from \(1.00\) (\(n{=}2\)) to \(0.55\) (\(n{=}5\)), \(0.20\) (\(n{=}10\)), and reaches \(0.00\) for \(n\ge 15\). Thus, the model usually preserves the overall chronology even when it misses the exact sequence: larger lists are typically off by a few swaps. The normalized Cayley distance increases roughly monotonically from \(0.125\) (\(n{=}5\)) to \(0.411\) (\(n{=}24\)), i.e.,
from \(\sim\!0.125\,(n{-}1)\) to \(\sim\!0.41\,(n{-}1)\) swaps.
For \(n{=}24\) this corresponds to about \(9\!-\!10\) adjacent swaps on average. Overall, when temporal gaps are wide and factual recall is controlled, GPT‑4.1 is globally broadly right but locally brittle: it retains the broad historical order yet rarely produces a perfect sequence once the list grows beyond ten items.

\begin{figure}[htbp]
  \centering
  \includegraphics[width=\linewidth]{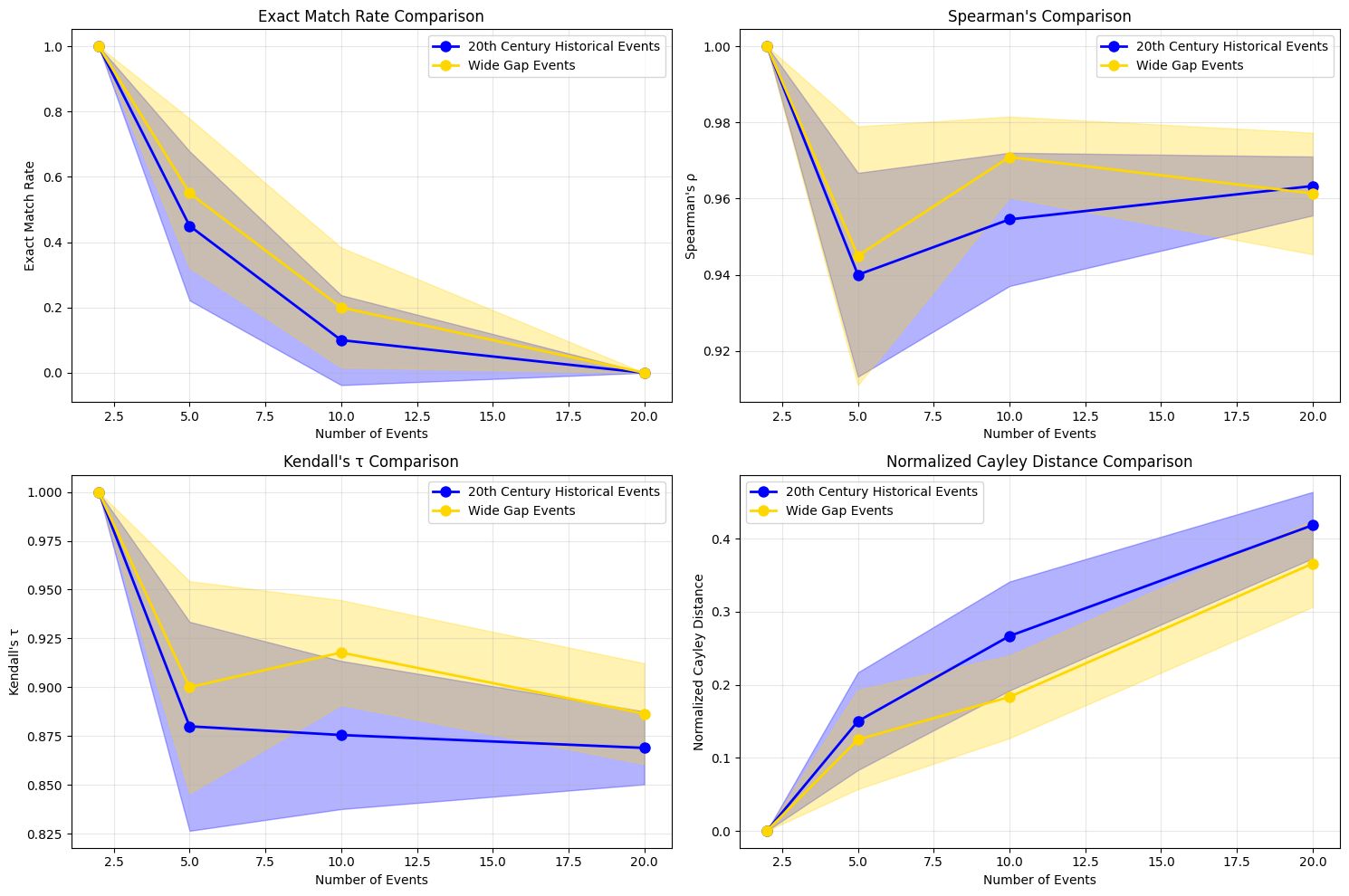}
  \caption{20th‑century (filtered) vs.\ wide time‑gap ordering.
  Lines show means over 20 trials; shaded bands denote \(\pm 2\) standard errors.}
  \label{fig:timescale-comparison}
\end{figure}

We further compare this result from our curated wide time-gap events with the one we obtained from \emph{20th‑century historical events} from a narrower timeline in Table \ref{tab:filtered-agg}. Figure~\ref{fig:timescale-comparison} juxtaposes their performances (2, 5, 10, and 20 items per trial).

\newpage

The comparison points to two conclusions:
\begin{itemize}[leftmargin=1.2em,nosep]
  \item \textbf{Wider time windows improve performance.} Ordering improves when events are separated by large temporal gaps (on the order of \(10^2\) years). Across list sizes, wide–time‑gap sets yield equal or higher ordering accuracy than same‑century baselines. This is what we would expect to see from human respondents --- it should be easier to sequence widely separated events --- but it is not obvious that the same should be true for LLMs.
  \item \textbf{Scaling list length worsens performance.} As the number of items increases, performance degrades for both—consistent with earlier ranking tasks—but less so under wide time gaps, with rank correlations still higher than their counterpart.
\end{itemize}

\section{Basic Sorting on U.S. Presidents} \label{sec:us-pres}

The historical events ordering tasks established a baseline for testing LLM understanding of chronology and allowed us to vary complexity through list length. In order to explore other dimensions of complexity, we turn now to a more structured dataset: the sequence of U.S. presidents. The U.S. has to date recorded 45 presidencies. However, Grover Cleveland and Donald Trump were elected to non-consecutive terms. To avoid potential ambiguities in sequencing presidents (e.g., whether should Trump be listed before or after Biden), we remove these two presidents from our list and work exclusively with the remaining 43.

\subsection{Methodology for the \texorpdfstring{U.\,S.\ Presidents}{US Presidents} Chronology Task}

For each list size \(n\in\{2,5,10,15,20,25,30,35,40,43\}\) we run \(20\) trials. In every trial we (i) apply knowledge verification—each candidate must return the correct presidency start year in a separate year query; (ii) sample \(n\) distinct presidents without replacement and uniformly shuffle them; (iii) prompt the model to order them chronologically; and (iv) score with our usual evaluation metrics. Full protocol and the workflow diagram appear in Appendix ~\ref{app:pres-method}.

\subsection{Findings \& Bottlenecks} \label{sec:us_pres_findings}
During our runs we observed that GPT‑4.1 occasionally omitted presidents that were present in the prompt or hallucinated additional names. (The hallucinated names were always presidents, just not presidents included in the sample.) In 42 of the 200 trials, the LLM omits at least one name from list in the prompt. The problem first appears at \(n{=}15\) (3/20 trials) and grows steadily, peaking at \(n{=}25\) and \(n{=}30\) where more than half the trials omit at least one name. 64 trials return presidents that did not appear in the prompt. Hallucinations are rare below \(n{=}25\) but dominate the longest lists, affecting 85 \% of \(n{=}40\) trials and 95\% of the full \(n{=}43\) trials, but there are no omissions or hallucinations for \(n\le10\).

From Table \ref{tab:who‑missing‑extra}, early‑period leaders such as James K.\ Polk and Andrew Jackson are most
often \emph{missing}; modern figures—Obama, Biden, both Bushes—are frequently \emph{added} even when absent from the prompt. This hallucination on a rather relatively simple task like sorting a subset of a well-known sequence (U.S. Presidents) may occur due to the model resorting to the full list it was being pre-trained on and ignoring the constraints given to only use the provided items. The pattern may also be similar to human errors in the sense that Polk is fairly obscure and recent presidents are much more salient.

To prevent omissions and hallucinations from distorting our earlier defined metrics, we need to adapt the evaluation pipeline  to account for such discrepancies. We introduced a dedicated cleaning stage before scoring each trial (as formalized in Appendix \ref{app:adapted-eval}).  First, every row whose \texttt{predicted\_rank} is \textsc{NaN} is removed, eliminating presidents that the model failed to mention.  Next, we discard rows whose \texttt{predicted\_rank} is numerically greater than or equal to \(n_{\text{prompt}}\); these correspond to hallucinated names that were not present in the input list.  If, after this pruning, no valid president–rank pairs remain, the trial is marked \emph{invalid} and omitted from aggregate statistics. We then compute metrics on the surviving pairs. Each trial also records auxiliary counters—\textsc{missing} for the number of omitted presidents, \textsc{extra} for hallucinated names, and \textsc{valid\_pairs} for the size of the filtered set—so that omission and hallucination rates can be reported alongside the main performance metrics.  These safeguards confine the evaluation to genuine ordering errors and shield the analysis from missing or superfluous names.


\begin{table}[htbp]
  \caption{Performance summaries}
  \label{tab:both}
  \centering
  \captionsetup[subtable]{position=top,skip=\baselineskip}

  \begin{subtable}[t]{0.70\linewidth}
    \caption{Per-size means ($\mu$) and standard errors (SE $=\sigma/\sqrt{20}$) after applying the cleaning pipeline.}
    \label{tab:summary-stats}
    \centering
    \makebox[\linewidth][c]{%
      \begin{tabular}{@{}c ccccccc@{}}
        \toprule
        \multirow{2}{*}{$n$} &
        \multicolumn{2}{c}{Exact match} &
        \multicolumn{2}{c}{Spearman's \(\rho\)} &
        \multicolumn{2}{c}{Kendall's \(\tau\)} &
        Cayley \\
        \cmidrule(lr){2-3}\cmidrule(lr){4-5}\cmidrule(lr){6-7}
         & $\mu$ & $\mathrm{SE}$ & $\mu$ & $\mathrm{SE}$ & $\mu$ & $\mathrm{SE}$ & $\mu$ \\
        \midrule
          2 & 0.96 & 0.04 & 0.998 & 0.00 & 0.997 & 0.00 & 0.00 \\
          5 & 0.87 & 0.08 & 0.973 & 0.02 & 0.960 & 0.03 & 0.20 \\
         10 & 0.36 & 0.11 & 0.961 & 0.02 & 0.932 & 0.02 & 1.72 \\
         15 & 0.20 & 0.09 & 0.960 & 0.01 & 0.927 & 0.02 & 3.27 \\
         20 & 0.10 & 0.07 & 0.971 & 0.01 & 0.929 & 0.01 & 6.20 \\
         25 & 0.09 & 0.08 & 0.967 & 0.01 & 0.930 & 0.01 & 6.90 \\
         30 & 0.00 & 0.00 & 0.963 & 0.01 & 0.948 & 0.02 & 6.60 \\
         35 & 0.00 & 0.00 & 0.992 & 0.00 & 0.986 & 0.01 & 3.65 \\
         40 & 0.00 & 0.00 & 0.993 & 0.00 & 0.989 & 0.01 & 3.10 \\
         43 & 0.00 & 0.00 & 1.000 & 0.00 & 1.000 & 0.00 & 0.05 \\
        \bottomrule
      \end{tabular}
    }
  \end{subtable}

  \medskip

  \begin{subtable}[t]{0.70\linewidth}
    \caption{Per-size means ($\mu$) and standard errors (SE $=\sigma/\sqrt{20}$) for error counts.}
    \label{tab:name-errors}
    \centering
    \makebox[\linewidth][c]{%
      \begin{tabular}{@{}c cc cc@{}}
        \toprule
        \multirow{2}{*}{$n$} &
        \multicolumn{2}{c}{missing names} &
        \multicolumn{2}{c}{extra names} \\
        \cmidrule(lr){2-3}\cmidrule(lr){4-5}
         & $\mu$ & $\mathrm{SE}$ & $\mu$ & $\mathrm{SE}$ \\
        \midrule
          2 & 0.00 & 0.00 & 0.60 & 0.28 \\
          5 & 0.00 & 0.00 & 0.00 & 0.00 \\
         10 & 0.00 & 0.00 & 1.00 & 0.46 \\
         15 & 0.20 & 0.09 & 0.07 & 0.06 \\
         20 & 0.30 & 0.16 & 0.00 & 0.00 \\
         25 & 0.70 & 0.16 & 0.25 & 0.20 \\
         30 & 1.05 & 0.30 & 0.55 & 0.36 \\
         35 & 0.90 & 0.28 & 2.50 & 0.68 \\
         40 & 0.35 & 0.22 & 3.25 & 0.42 \\
         43 & 0.05 & 0.05 & 1.65 & 0.13 \\
        \bottomrule
      \end{tabular}
    }
  \end{subtable}
\end{table}
\paragraph{Summary statistics after cleaning}

Table \ref{tab:summary-stats} and \ref{tab:name-errors} report the mean and standard deviation of all evaluation metrics, averaged over \(20\) trials for each list size.  The columns \textsc{missing\_names} and \textsc{extra\_names} quantify the average number of presidents that were omitted or hallucinated \emph{before} cleaning.  The most striking feature of Table \ref{tab:summary-stats}, illustrated in Figure~\ref{fig:ushape-curve} (Appendix \ref{app:gpt-ordering-error}), is the U-shaped pattern for the correlation measures: they generally fall as the list length increases but then rebound as $n$ increases toward the complete list. This pattern suggests that the LLM ``knows" the complete timeline of U.S. presidents and has more difficulty ordering a random subset. However, as we saw with the timeline of historical events, the exact match rate drops sharply as the list length grows. 

We also summarize per-position misplacement with the mean absolute rank difference (MARD), illustrated in Figure~\ref{fig:MAE-position}. MARD per position stays modest for the positions $r\le20$, then climbs steeply, reaching its peak around $r\approx 30\text{–}35$. These high‑error slots coincide with the mid‑19\textsuperscript{th}-century presidents most often omitted or replaced. In Table~\ref{tab:century-accuracy}, we group presidents by century to confirm this trend: accuracy is highest for well‑known 18th‑century figures and degrades through the 19th and 20th centuries, before improving for the still‑evolving 21st‑century cohort. All in all, these findings suggest that GPT‑4.1’s ordering reliability depends jointly on both the list length and the historical salience of the individual names.

\newpage

\subsection{Experimenting with Large Reasoning Models (LRMs)}

We also tested chronological ordering of U.S. presidents with a few large reasoning models, namely Claude 3.7 Sonnet and the newly-released GPT-5 with varying reasoning efforts (minimal, low, medium, and high). In this section, we discuss how they fare compared with the normal LLMs and provide working hypotheses to explain the results. 

Recent “large reasoning models” such as OpenAI's ChatGPT-o3 or Anthropic's Claude 3.7 Sonnet add an explicit, controllable deliberation phase to the usual input→output loop. To test whether this explicit “scratch‑pad” reasoning improves ordering outcomes, we test the president-ordering task on a large reasoning model (LRM) with \emph{extended thinking} support.\footnote{Anthropic’s developer docs describe extended thinking and streaming: \url{https://docs.anthropic.com/en/docs/build-with-claude/extended-thinking} and \url{https://docs.anthropic.com/en/docs/build-with-claude/messages/streaming}.} With \emph{extended thinking}, the model can allocate a separate budget of “thinking” tokens to produce intermediate reasoning before emitting the final answer. Those thinking tokens are part of the normal token accounting (they count toward the context window and are billed as output), and the budget is developer‑tunable (via budget tokens). Extended thinking can also be returned to the caller for inspection, which allows us to study how much (and when) extra reasoning helps chronological ordering, making it a good testbed for comparing standard vs. extended reasoning on the same prompts for our analysis. 

We also evaluated OpenAI’s GPT-5, which differs from Claude 3.7 Sonnet in two ways: (i) it is a unified reasoning model that automatically \emph{routes} between fast replies and deeper thinking, and (ii) it allows us to tune a \texttt{reasoning\_effort} hyperparameter (e.g., minimal/low/medium/high) that controls internal reasoning tokens but does \emph{not} return an inspectable reasoning stream. We include GPT-5 to test whether tunable test-time reasoning improves chronological ordering even without visible Chain-of-Thought traces \citep{OpenAI2025GPT5,OpenAI2025GPT5Devs,OpenAI2025Reasoning,OpenAI2025PromptingGuide}.

\paragraph{Experimental design}
We run head‑to‑head controlled comparisons (identical prompt temperature, sample\_sizes, and trials per sample\_sizes) with the same president‑ordering methodology outlined earlier and in Appendix \ref{app:pres-method}  tasks under eight regimes: GPT-4.1 (results directly from Section \ref{sec:us_pres_findings}), Claude 3.7 Sonnet without Extended Thinking (ET), Claude 3.7 Sonnet with Extended Thinking (ET), GPT-5 (reasoning\_effort = minimal), GPT-5 (reasoning\_effort = low), GPT-5 (reasoning\_effort = medium), GPT-5 (reasoning\_effort = high), and the non-reasoning GPT-5 (latest) variant as a control.

\begin{table}[htbp]
  \centering
  \footnotesize
  \caption{Claude Sonnet~3.7 extended‑thinking run‑time parameters.}
  \begin{tabular}{@{}ll@{}}
    \toprule
    Parameter & Value \\ \midrule
    Temperature & \(T{=}1.0\) (ET runs); \(T{=}0.0\) (non‑ET baselines) \\
    Thinking budget & \texttt{budget\_tokens} \(= 5000\) \\
    Max response tokens & \texttt{max\_tokens} \(= 8000\) \\
    Rationale & 5k budget avoids ET truncation; 8k headroom for final text/tool I/O \\
    \bottomrule
    \label{fig:et-token-flow}
  \end{tabular}
\end{table}

All models receive identical prompts and shuffled name lists at each list length \(n\in\{2,5,10,15,20,25,30,35,40,43\}\).
For ET we follow the provider’s recommended usage: temperature \(T{=}1.0\), \texttt{budget\_tokens}\(=5000\) for the \emph{thinking} stream, and \texttt{max\_tokens}\(=8000\) for the visible response. The 5k thinking budget was chosen to avoid truncation of the hidden \emph{thinking} stream in our prompts, while \(\texttt{max\_tokens}{=}8000\) comfortably accommodates the visible answer plus any tool‑usage text. All non‑ET baselines were run at \(T{=}0\). \footnote{ET requires \(T{=}1.0\) and manages a separate hidden ``thinking'' stream; see Table.~\ref{fig:et-token-flow} and the model documentation.}

\subsubsection{Key findings}

\begin{figure}[htbp]
  \centering
  \includegraphics[width=\linewidth]{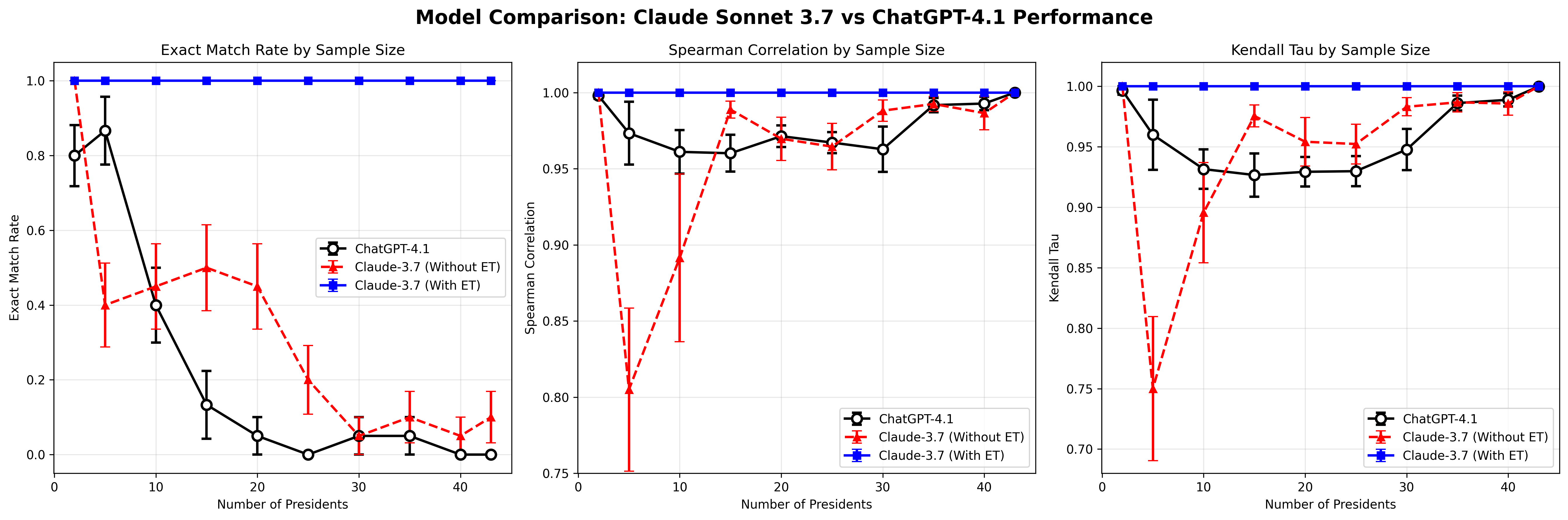}
  \caption{\textit{Claude 3.7 with Extended Thinking} (blue) dominates—achieving \(100\%\) Exact match at all \(n\)
Points show trial means; error bars are \(\pm 2\) s.e.; small horizontal jitter prevents overlap.}
  \label{fig:model-compare}
\end{figure}

\paragraph{Claude 3.7 Sonnet (standard) vs.\ GPT‑4.1}

To our surprise, Claude~3.7 with ET achieves \emph{perfect chronological ordering} across all list sizes (Fig.~\ref{fig:claude-et-panels}): exact‑match \(=1.00\) at every \(n\), and normalized Cayley distance indistinguishable from zero. In sharp contrast, Claude~3.7 \emph{without} ET and GPT‑4.1 reproduce the characteristic pattern from earlier task: rank correlations remain high, but the exact‑match rate collapses as \(n\) grows (Fig.~\ref{fig:model-compare}). The improvement with ET is not a small calibration gain; it is a qualitative shift from “near‑correct orderings with frequent small inversions” to \emph{error‑free} outputs. Panels in Fig.~\ref{fig:claude-et-panels} (bottom row) also show that ET suppresses missing or extra hallucinated outputs completely. Without ET, runs often contain \textsc{missing}, and more so \textsc{extra} names (length mismatches) even when rank correlations are high; with ET, these errors are essentially absent. ET flattens curves on all metrics: performance is invariant to \(n\) within our tested range, indicating that the model’s hidden reasoning is sufficient to maintain global consistency over long lists.

A striking secondary result is that \emph{without} extended thinking (ET), Claude~3.7 does not reliably outperform GPT‑4.1 on this benchmark despite it being a large reasoning model which we believed would perform better than GPT-4.1; in fact, at \(n{=}5\) Claude lags 4.1 (Fig.~\ref{fig:model-compare}). 

\paragraph{Why does Extended Thinking help? A working hypothesis}

We do not know the mechanism within the model that allows for such flawless performance. However, we hypothesize that because ET grants the model a large, private scratchpad to deliberate before emitting any visible text, this may enable a reliable internal routine: (1) enumerate all presidents; (2) verify membership against the criterion implicit in the task; (3) perform pairwise chronology checks; (4) remove any duplicates; and (5) emit the final ordered list.

Separating \emph{thinking} from \emph{outputting} may reduce early‑commitment and exposure‑bias errors, and it allows multiple self‑checks without incurring user‑visible tokens. We give an excerpt from Claude 3.7 Sonnet with ET enabled, taken from a trial that required ordering a subset (30) of presidents in Appendix \ref{app:et-trace}. This thinking trace gives us useful information to make an inference on the reasoning model's exceptional ordering capability. The model first reconstructs an external scaffold (terms in office), which supplies a consistent comparison key. It then explicitly marks incorrect names (``not on the list''), showing evidence of a \emph{set‑membership} check before ordering. The final output is a clean, deduplicated sequence without dates—consistent with the output format requested in the prompt —indicating a separation between computation and presentation. Together these behaviours align with our hypothesis: ET supplies the private working memory needed to complete filtering and ordering reliably before committing to the final output.

\begin{figure}[htbp]
  \centering
  \includegraphics[width=0.70\linewidth]{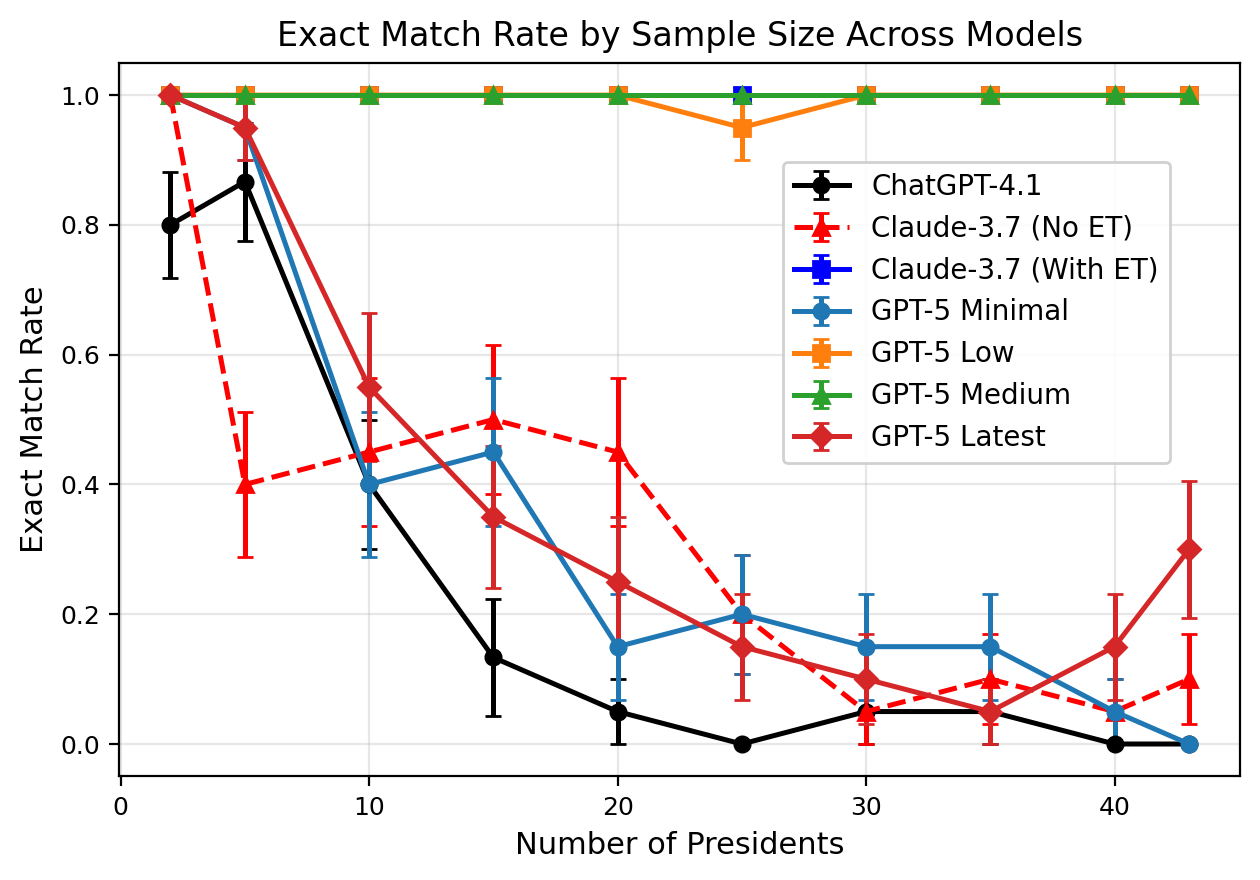}
  \caption{Exact match by list size. \textbf{GPT-5 (medium, high)} and \textbf{Claude 3.7 with Extended Thinking} achieve \emph{flawless} Exact match (\(100\%\)) across all \(n\). \textbf{GPT-5 low} is near-perfect, with only a minor dip at mid sizes. The remaining models—\textbf{GPT-5 minimal}, \textbf{Claude 3.7 without ET}, and \textbf{GPT-4.1}—are broadly comparable to one another and lose Exact match rapidly as \(n\) increases. Points show trial means; error bars are \(\pm 2\) s.e.; small horizontal jitter prevents overlap.}
  \label{fig:exact-match-model-comparison}
\end{figure}

\paragraph{GPT-5 with varying reasoning efforts}
Meanwhile, GPT-5 shows a sharp effort–accuracy transition. With \emph{medium} and \emph{high} reasoning effort the model achieves \emph{perfect} chronological ordering at all list sizes (Exact match \(=1.00\); \(\rho=\tau=1.00\). The \emph{low} setting is near-perfect, with a single dip at \(n{=}25\) (EM \(=0.95\)); otherwise performance is indistinguishable from perfect. In contrast, the \emph{minimal} and \emph{no-reasoning} variants mirror non-reasoning LLMs: rank correlations remain very high (\(\rho,\tau\ge 0.94\)) but exact match collapses as list size increases. We observe a clear pattern: as GPT-5’s reasoning effort increases from minimal/low to medium/high, Exact match significantly improves and reaches 100\% at medium/high.  This supports our hypothesis that giving the model more test-time reasoning budget (i.e., more “space”/inference time/private scratchpad) enables it to excel at chronological orderings by providing it with a simple internal "checklist": check membership, compare dates, then produce a single consistent order—to complete. 

The effect mirrors what we see with Claude 3.7’s Extended Thinking, where an explicit thinking stream plays a similar role. We treat both systems as black boxes, but the alignment between effort/ET and the accuracy gains suggests that added deliberation is the main driver for such a dramatic increase in performance metrics.

\section{Conditional Sorting} 

The experiments reported thus far have each evaluated a single, direct form of chronological understanding: present a shuffled list of events and ask the model to restore the correct order. We now turn to tasks in which the events to be ordered are defined implicitly through some condition (e.g., “presidents born on a Monday”) rather than sampled randomly and presented explicitly. This allows us to explore aspects of task complexity that differ from list length.

Requiring the LLM to filter the events that satisfy some condition adds to the complexity of the task. However, the reasoning it requires could potentially act as a scaffold that provides the LLM with greater insight into the selected names (compared to being presented with a random list), helping it avoid errors. Our experiments will test these competing hypotheses and conclude that the added complexity dominates.

\subsection{Methodology for the \texorpdfstring{U.\,S.\ Presidents}{US Presidents} One-Prompt Conditional Sorting Task}
\label{sec:thinking-method}

The direct--ordering experiments of U.S. presidents in the earlier section showed that GPT--4.1’s exact match rate collapses completely once the list exceeds about ten names. We, therefore, naturally conjecture that a \emph{two‑step} prompt---urging the model to \emph{think} by first \emph{filtering} the candidate set before ordering it---may nudge the LLM into a more deliberative “System‑2’’ reasoning and thus lower its error rate. Our central question is straightforward: does GPT‑4.1 sort more accurately when it \emph{discovers} the relevant names itself, or when it is simply \emph{given} the correct subset and asked to order it? 

Each trial begins with a single shuffled list of 43 unique presidents like before. Two matched conditions are then run on that \emph{same} shuffled order. In the \emph{self‑filtering–sorting} condition, the model is told to filter for a factual criteria \(c\) and then to sort the surviving names chronologically by presidency. In the \emph{given‑names sorting} condition, the model receives exactly the names that satisfy \(c\)---in the same relative order they appeared in the full list---and is asked only to sort them in a separate API call.

We experimented with various criteria:
the state in which each president was born—\textsc{Ohio} (7 names), \textsc{Virginia} (8 names), \textsc{Massachusetts} (4 names), and whether their birth year was even, \textsc{EvenYears} (22 names)—and found that GPT‑4.1 almost never filtered perfectly (single‑digit success rates over 100 trials). To balance tractability and difficulty, we adopted the combined criteria to filter the presidents being born in either \textsc{OhioOrVirginia} (15 names): large enough to make ordering non‑trivial, but small enough that perfect filtering occurs with reasonable frequency. In short, it is “just right’’ for our contrast. 

\paragraph{Sharp contrast protocol}
To make a clean comparison, we retain \emph{only} those self‑filtering runs in which the model’s filtered set $G_c^{(t)}$ exactly matches the ground truth $G_c$.
If the final list is missing any required name or contains any extraneous name, the filtering step is marked \emph{invalid} and that trial is discarded from the ordering comparison (duplicates are ignored for filtering, but counted as an ordering error later).  Each trial begins with a single shuffled list of all 43 unique presidents (Grover Cleveland and Donald Trump removed). Two matched conditions are then run on that \emph{same} shuffle:

\begin{description}[nosep,leftmargin=1.6em,style=nextline]
  \item[\textsc{self\_filtering\_sorting}] The model receives the full list and a criteria \(c\) (e.g., ``born in OhioOrVirginia''), filters, then orders the survivors by presidency.
  \item[\textsc{given\_names\_sorting}] The model receives exactly the \(G_c\) names—presented in the same relative order as in the full list—and is asked only to sort them.
\end{description}

We give an illustrative example here to illustrate the algorithm. Let the shuffled list be \([F,B,D,G,H,E,C,A]\) and \(c=\) ``born in OhioOrVirginia'', giving \(G_c=\{B,E,G\}\).  
\begin{enumerate}[nosep,leftmargin=1.5em]
  \item \textsc{self\_filtering\_sorting}: Input the full list; instruct “filter for OhioOrVirginia, then order chronologically.” Correct behavior: output \([B,G,E]\) \(\rightarrow\) reorder to \([B,E,G]\).
  \item \textsc{given\_names\_sorting}: Input only \([B,G,E]\); instruct “order chronologically.” Output \([B,E,G]\).
\end{enumerate}

The full conditional sorting protocol formalization is present in Appendix \ref{app:conditional-protocol}. For each criterion we report: (i) the perfect‑filter pass rate, (ii) mean/SD of each ordering metric in both conditions, and (iii) paired deltas.

\begin{figure}[t]
  \centering
  \includegraphics[width=1.00\linewidth]{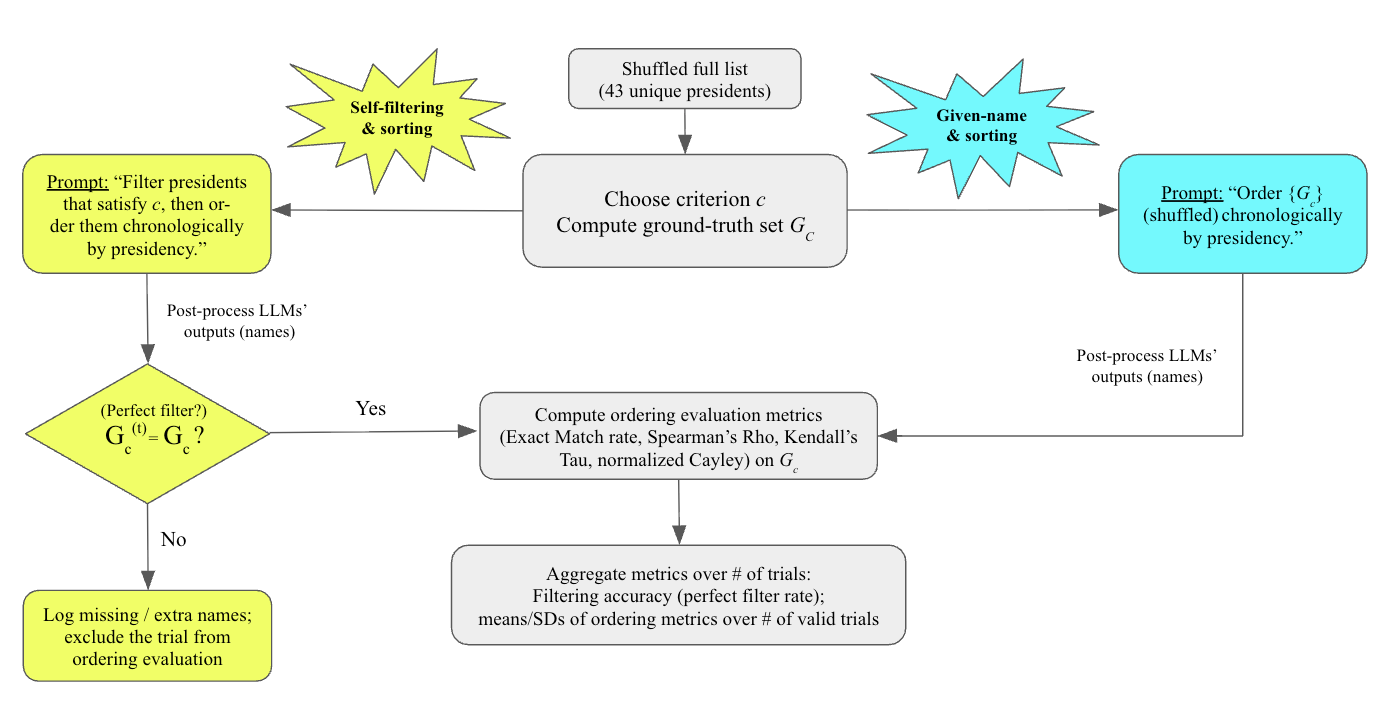}
  \caption{%
    Conditional sorting pipeline: self-filtering vs.\ given-names. Only trials with
    ${G}_c^{(t)} = G_c$ on the left enter the paired comparison.
    Duplicates count toward ordering errors but not the filtering decision.
  }
  \label{fig:week4-flow}
\end{figure}

\subsection{Key findings}

\subsubsection{Filtering is the dominant failure mode}

Alarmingly, for the criteria $c=\textsc{FirstNamesStartingWithABorC}$ (ground‑truth size $n_{\mathrm{gt}}=\lvert G_c\rvert=8$), the model achieved only a whopping 2\% perfect filtering: in $2/100$ runs the predicted set $\hat G_c$ exactly matched $G_c$; consequently there were too few valid runs to support a meaningful ordering analysis in the self–filter condition. In this task, GPT-4.1 also hallucinates erroneous names. All five have surnames beginning with “B,” consistent with a last‑name‑initial confusion. (Counts are out of 100 trials.) Interestingly, the most common false inclusions are modern presidents whose \emph{surnames} begin with B (e.g., \emph{B}iden, \emph{B}ush, \emph{B}uchanan, Van \emph{B}uren), indicating that the model often applied the initial to the last name rather than the first. We also observe occasional nickname/alias leakage (e.g., “Jimmy” vs.\ “James” Carter). The error profiles show a strongly \emph{asymmetric} failure mode: the model tends to overselect rather than underselect presidents. Across trials it frequently appends presidents who do not satisfy the A/B/C first‑name criterion, while true positives are seldom omitted; even in runs that contain both misses and extras, the extras typically dominate.

We observed the same trend for \textsc{OhioOrVirginia} filter; the self‑filtering pass rate was essentially \emph{zero}. Out of \(100\) self‑filtering trials, not a single run achieved perfect set equality \(\hat G_c^{(t)} = G_c\), leaving \emph{no} valid trials from which to estimate ordering accuracy under self‑filtering. The most frequently \emph{omitted} ground‑truth name \(G_c\setminus\hat G_c^{(t)}\) was Woodrow Wilson (born in Staunton, Virginia), missed in \(28/100\) trials.  The most common extra inclusions \(\hat G_c^{(t)}\setminus G_c\) in \(100/100\) trials were James Buchanan and Millard Fillmore (neither born in OhioOrVirginia). 

Taken together, the results demonstrate that when filtering and ordering are bundled into a single instruction, GPT‑4.1 fails at the \emph{filtering} stage entirely, leaving too few valid runs for meaningful ordering comparisons across the self-filtered and given-names tasks.

\subsection{Conditional Sorting with Large Reasoning Models}

After seeing Claude 3.7 Sonnet with Extended Thinking's infallible performance on the ordering task, we re-ran the conditional-sorting task with reasoning models, in the hope that they would also show improved accuracy on the more difficult filtering step so ordering performance analysis can be done.

Specifically, we test: (i) \textbf{Claude 3.7 Sonnet with and without Extended Thinking}, and (ii) \textbf{GPT-5 at \emph{medium} reasoning effort}, which in our earlier U.S. Presidents ordering experiments closely matched Claude 3.7 Sonnet with Extended Thinking performance and thus serves as a natural default comparison. Both are evaluated alongside the GPT-4.1 baseline under identical prompts and list sizes. If filtering failures are partly due to premature commitments or limited reasoning during inference time, our hypothesis is that reasoning models with ET and its equivalent may reduce such errors.

We used the same identical prompts, shuffled input list per trial to allow paired comparisons. We vary only the ET switch and the sampling settings required by Claude (same as Table \ref{fig:claude-et-panels}). All other parameters (stop sequences, tool use off, etc.) are held fixed. We ran four condition–criterion cells, each with \(100\) independent trials.

\subsubsection{Key findings}

\begin{table}[htbp]
\centering
\caption{Filtering accuracy across models and conditions (ET = Extended Thinking).}
\label{tab:filtering-accuracy}
\begin{tabular}{llcccc}
\toprule
\textbf{Model} & \textbf{Condition} & \textbf{Filtering Accuracy} & \textbf{n\_correct} & \textbf{n\_total} & \textbf{ET?} \\
\midrule
Claude 3.7 Sonnet & ABC First names       & 0.81 & 81 & 100 & No \\
Claude 3.7 Sonnet & ABC First names       & 0.99 & 99 & 100 & Yes \\
Claude 3.7 Sonnet & OhioOrVirginia   & 0.02 &  2 & 100 & No \\
Claude 3.7 Sonnet & OhioOrVirginia   & 0.98 & 98 & 100 & Yes \\
GPT‑4.1    & ABC First names       & 0.02   & 2 & 100 & N/A \\
GPT‑4.1    & OhioOrVirginia   & 0.00   & 0 & 100 & N/A \\
GPT-5 (medium)     & ABC First names       & 1.00   & 100 & 100 & N/A \\
GPT-5 (medium)      & OhioOrVirginia   & 1.00   & 100 & 100 & N/A \\
\bottomrule
\end{tabular}
\end{table}

\begin{table}[htbp]
\centering
\caption{Claude 3.7 Sonnet (Extended Thinking) ordering performance across metrics.}
\label{tab:ordering-metrics}
\begin{tabular}{llcccc}
\toprule
\textbf{Task} & \textbf{Metric} & \textbf{Value} & \textbf{$n_{\text{correct}}$} & \textbf{$n_{\text{total}}$} & \textbf{Condition} \\
\midrule
ABC First names & Avg.\ Spearman's $\rho$ & 0.999 & 100 & 100 & Given Names \\
ABC First names & Avg.\ Kendall's $\tau$  & 0.999 &  99 & 100 & Given Names \\
ABC First names & Exact Match Rate        & 0.99  &  99 & 100 & Given Names \\
ABC First names & Avg.\ Spearman's $\rho$ & 1.000 & 100 & 100 & Self-Filtered \\
ABC First names & Avg.\ Kendall's $\tau$  & 1.000 & 100 & 100 & Self-Filtered \\
ABC First names & Exact Match Rate        & 0.99  &  99 & 100 & Self-Filtered \\
\midrule
OhioOrVirginia  & Avg.\ Spearman's $\rho$ & 0.997 &  99 & 100 & Given Names \\
OhioOrVirginia  & Avg.\ Kendall's $\tau$  & 0.995 &  99 & 100 & Given Names \\
OhioOrVirginia  & Exact Match Rate        & 0.82  &  82 & 100 & Given Names \\
OhioOrVirginia  & Avg.\ Spearman's $\rho$ & 1.000 & 100 & 100 & Self-Filtered \\
OhioOrVirginia  & Avg.\ Kendall's $\tau$  & 1.000 & 100 & 100 & Self-Filtered \\
OhioOrVirginia  & Exact Match Rate        & 0.97  &  97 & 100 & Self-Filtered \\
\bottomrule
\end{tabular}
\end{table}

As expected, Table \ref{tab:filtering-accuracy} shows that reasoning models outperform GPT-4.1 in the filtering step. Notably, GPT-5 (medium) achieved \emph{perfect} filtering accuracy on both criteria. On those knowledge-verified
subsets, its downstream ordering was also perfect in our runs. Claude 3.7 Sonnet also edges out GPT-4.1 in the filtering step, consistent with our earlier hypothesis that LRMs might be better for conditional sorting. Without ET, \textsc{FirstNamesStartingWithABorC} criteria (string‑based) is far easier than \textsc{OhioOrVirginia} criteria (fact-based) for the reasoning model.

More interestingly, Extended Thinking transforms Claude 3.7’s \emph{filtering} accuracy from inconsistent to near‑perfect. On the harder factual criteria \textsc{OhioOrVirginia}, accuracy jumps from 0.02 without ET to 0.98 with ET. Even on the easier lexical criteria (\textsc{FirstNamesStartingWithABorC}), ET raises accuracy from 0.81 to 0.99. With ET, the gap largely disappears, indicating the model can execute factual checks once it is allowed to deliberate. ET also gives Claude a private scratchpad to enumerate candidates and verify membership, which sharply reduces both omissions and over‑inclusion of presidential names. By contrast, GPT‑4.1—without an ET mechanism—performs the worst on both criteria.

With Claude 3.7 Sonnet and ET, enough \emph{perfectly filtered} trials now exist to make downstream ordering comparisons meaningful which tackles the bottlenecks we encountered with GPT‑4.1 where filtering yields too few valid trials, biasing ordering metrics upward and limiting interpretability.

\begin{figure}[htbp]
  \centering
  \captionsetup[subfigure]{font=small,justification=raggedright,singlelinecheck=false}

  \begin{subfigure}[t]{0.48\linewidth}
    \centering
    \includegraphics[width=\linewidth]{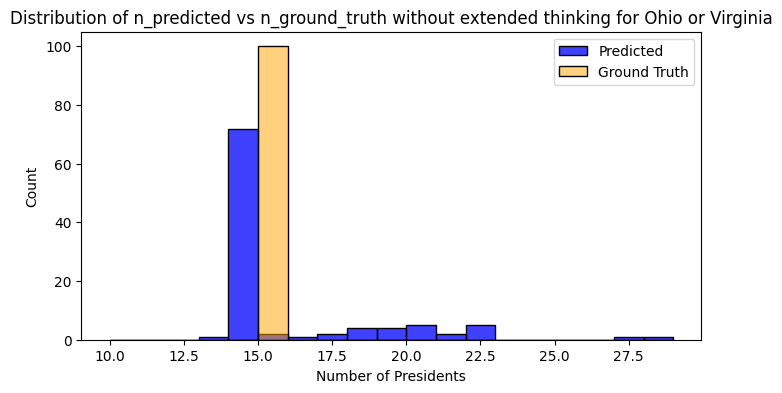}
    \caption{Without ET. Predicted list sizes (blue) are widely spread (around 14--29), while the ground-truth size (orange) spikes at 15.}
  \end{subfigure}\hfill
  \begin{subfigure}[t]{0.48\linewidth}
    \centering
    \includegraphics[width=\linewidth]{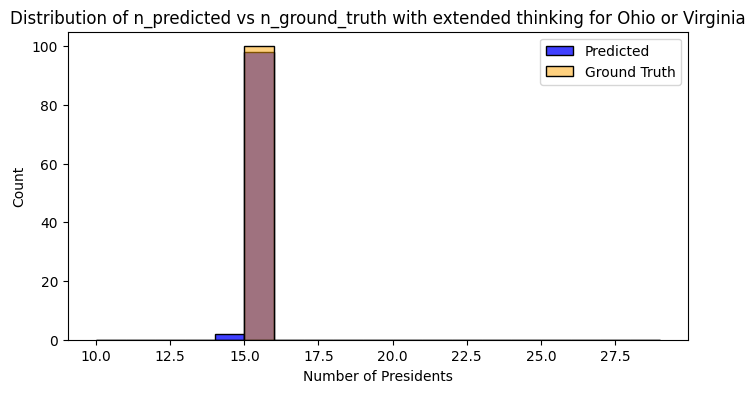}
    \caption{With ET. Predicted list sizes collapse to the correct value (15) in almost all trials.}
  \end{subfigure}

  \caption{\textbf{Claude 3.7 Sonnet, self-filtering on \texorpdfstring{$\text{Ohio}\lor\text{Virginia}$}{OhioOrVirginia} (100 trials).}
  Histograms compare the number of presidents output by the model (blue) with the ground truth (orange). Extended Thinking (ET) removes the over/under-selection seen without ET.}
  \label{fig:ov-length-dist-et}
\end{figure}

In terms of rank correlation metrics, Extended Thinking also improved both metrics up to around 1.0 for every group (\textsc{FirstNamesStartingWithABorC} and \textsc{OhioOrVirginia}; given‑names and self‑filtered). Without ET, \textsc{FirstNamesStartingWithABorC} self‑filtered group performance drops sharply, and \textsc{OhioOrVirginia} given‑names performance is decent but not perfect. The most striking result lies in the exact match rate: Extended Thinking lifts exact match close to 1.0 across the board. Without ET, exact match is near zero for self‑filtered \textsc{FirstNamesStartingWithABorC} and\textsc{OhioOrVirginia}, and only around 0.37 for \textsc{FirstNamesStartingWithABorC} given‑names. Overall, we believe the greatest analogy to explain this is that ET converts a brittle “filter‑then‑sort while talking” problem into a reliable “think privately, verify, then speak” workflow—reducing filtering errors and ordering slips.

\paragraph{Per Position Accuracy Error Analysis}
Just like before, we analyzed per-position accuracy for Claude 3.7 Sonnet among the given name trials without length mismatch (pure ordering errors) for each task type and criteria (we focus only on given name tasks with no ET since ET almost got everything perfectly right -- we do not know how informative the error analysis is going to be.) Self-filtered tasks are also not the most informative to analyze here since despite improving filtering accuracy, it is still low for per-position analysis, especially when most errors are attributed to length mismatches instead of pure ordering errors.

\begin{figure}[t]
  \centering
  \begin{subfigure}[t]{0.48\linewidth}
    \centering
    \includegraphics[width=\linewidth]{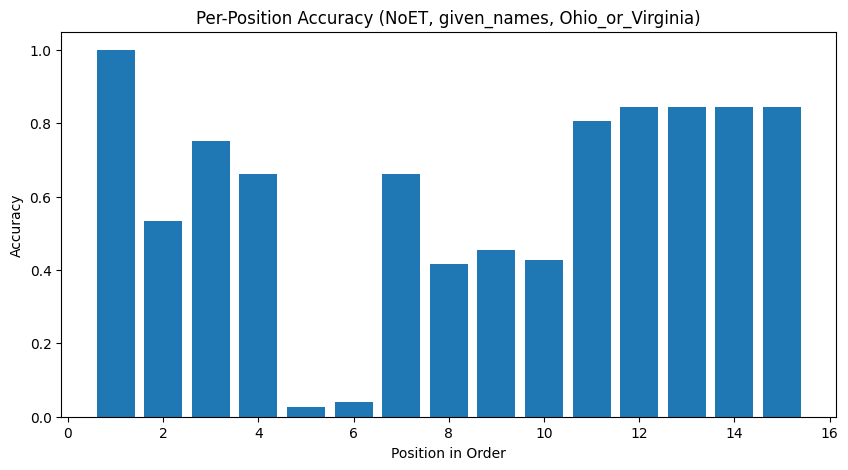}
    \caption{Ohio$\lor$Virginia (NoET, given‑names). Among 77 trials with no
    length mismatch, \textbf{2} (2.6\%) were perfect. Accuracy is 1.00 at the
    head, dips in the mid‑list, then climbs to $\geq$0.80 for the tail.}
    \label{fig:pos-acc-ov-noet}
  \end{subfigure}\hfill
  \begin{subfigure}[t]{0.48\linewidth}
    \centering
    \includegraphics[width=\linewidth]{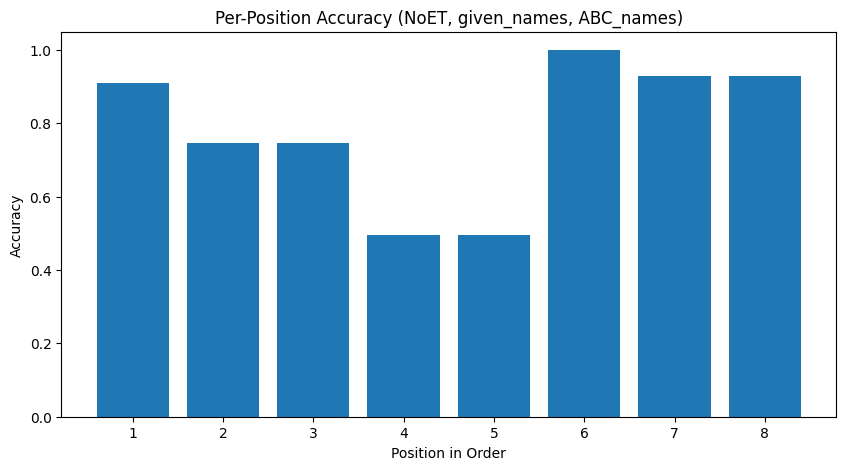}
    \caption{\textsc{FirstNamesStartingWithABorC} filter (NoET, given‑names). Among 99 trials with no length
    mismatch, \textbf{37} (37.4\%) were perfect. Accuracy is generally high with
    a small mid‑list dip.}
    \label{fig:pos-acc-abc-noet}
  \end{subfigure}
  \caption{\textbf{Per‑position accuracy without extended thinking (ET) in the
  \emph{given‑names} setting.} Bars show, for each rank, the fraction of trials
  placing the correct president at that position. These panels condition on
  equal list lengths, so errors reflect pure ordering rather than spurious
  insertions/deletions.}
  \label{fig:pos-acc-noet-given}
\end{figure}

Figure \ref{fig:pos-acc-noet-given} shows us a clear trend of the model excelling at ordering names in the beginning of the list, with degrading performance at the middle of the list and better performance again at the end of the list. This U‑shape is apparent in both criteria which suggests that the model may confuse the densely packed mid‑century presidencies even when the set is correct. The model performed worse mid-list for Ohio\(\lor\)Virginia as compared to \textsc{FirstNamesStartingWithABorC} names criteria; mid‑list placements are the most error‑prone when the criterion yields a broad, era‑spanning set with more names, whereas the smaller \textsc{FirstNamesStartingWithABorC} set is uniformly easier once names are provided. Another hypothesis is that stronger anchors exist at the beginning and the end of the list. The earliest and latest figures in the subset are historically salient and far apart in time, so the model locks them in due to primacy/recency anchoring, giving 1.00 accuracy at the head and around 0.80 at the tail.

\paragraph{Conclusion}
We have seen that across president–ordering and conditional–sorting benchmarks, Claude 3.7 Sonnet with \emph{extended thinking} (ET) and GPT-5 (medium) consistently outperforms both its own non‑ET variant and GPT‑4.1 on the axes that matter most for this task—strict set accuracy and exact chronological order. ET and reasoning modes in GPT-5 remove the filtering bottleneck in the conditional sorting tasks, driving filtering and ordering accuracy to near‑perfect. Without ET, Claude's performance is similar to GPT‑4.1 and even trails it at \(n{=}5\). 

With ET enabled, the two pipelines are essentially equivalent on ordering, and in some cases the self-filtered condition is slightly \emph{better} on the harder criteria (Table \ref{tab:ordering-metrics}). For Claude 3.7 Sonnet with Extended Thinking, we see near-identical exact-match on \textsc{FirstNamesStartingWithABorC} (0.99 vs. 0.99) and a small but consistent edge for self-filtered condition on the \textsc{OhioOrVirginia} criterion (0.97 vs. 0.82). More interestingly, GPT-5 with medium reasoning effort is the strongest model as it is able to achieve both perfect filtering and ordering ability, edging out Claude 3.7 Sonnet even with Extended Thinking. In short, conditional sorting mainly adds complexity via the filtering bottleneck; once that is overcome, it does slightly boost ordering performance—if anything, the self-filtered pipeline yields equal or slightly higher ordering accuracy than the given-names condition. Conditional sorting might propel the model to internally check and verify the candidate set through its private 'scratchpad' before ordering which helps boost performance, whereas given-names skips this construction.

\section{Anachronism detection}

Building on the earlier sections---where we attempted to abstract out LLMs' innate understanding of chronology---anachronism detection raises the bar. Instead of asking the LLM to order events or historical figures chronologically, we ask it to reliably judge whether it would be \emph{chronologically possible} for an event to occur. We ask, for example, whether it would have been chronologically possible for president A to meet president B, or whether A could have used a certain technological invention during their presidency. This requires the model to retrieve, compose, and comprehend overlapping timelines: the president's term (or lifespan), and the availability window of a technology, institution, or the duration of an event (in terms of first possible date and last possible date if any). Such a judgment hinges on the intersection of those intervals, instead of just a single sorted list of names. The task also contains added difficulties as merely regurgitating canonical presidential order or historical-event timelines from the LLM's training data will not be enough to excel; the model must align two (or more) independent timelines and spot any contradictions or inconsistencies. Here it must integrate world knowledge to render a binary judgment--possible vs. impossible. By moving from ordinal placement to feasibility detection, anachronism detection offers a measure of whether LLMs inherently possess an understanding of chronology rather than surface-level recall. Thus, it is the natural next step after our three ordering experiments.

\subsection{Experimental Design}

We conducted two anachronism experiments with increasing temporal complexities:

\begin{enumerate}[nosep,leftmargin=1.25em]
  \item \textbf{Single–boundary feasibility (first–possible date).}
        Each event \(e\) has a well-defined first introduction/availability year \(t_{\min}(e)\). A pair of (president, event) \((p,e)\) is labeled ``possible'' if and only if the president’s term window \([a(p),b(p)]\) intersects \([t_{\min}(e),\infty)\).
        \newline
  \item \textbf{Multi-timeline overlap.}
        Queries such as “were [n specific presidents] \(p_1,p_2,...,p_n\) all alive at the same time?” require checking \(\bigcap_{k}[\,\ell(p_k),d(p_k)\,]\neq\varnothing\),
        where \(\ell(\cdot),d(\cdot)\) are birth/death years. This directly tests LLMs' reasoning over multiple, partially overlapping intervals.
\end{enumerate}

Anachronisms can be understood as non-overlapping intervals in the Allen relations studied in \citet{chronosense}, but there are important differences between our tests and those in \citet{chronosense}. We do not probe individual Allen relations but rather require the LLM to determine whether any of the relations that would make an event possible actually hold. Also, our multi-timeline tests require an LLM to check for the intersection of multiple intervals, rather than just two.

\paragraph{Establishing groundtruth and removing grey zones}
For our first variant, we first established the groundtruth president-event pairs dataset by curating \(12\)–\(15\) activities with 
that only became possible after some date
(e.g., “fly in an airplane,” “make a telephone call,” “use generative AI tools”).

For each event/activity \(e\) (e.g., ``Flew in an airplane while president''), we record a \emph{first–possible year} \(t_{\min}(e)\) as the year of the technology’s invention or first public release date. Some events may also have a \emph{last–possible year} \(t_{\max}(e)\); but for events like invention with no last-possible year, we take \(t_{\max}(e)=+\infty\).

Many technologies exhibit a lag between invention and reliable in-office or commercialized use. To avoid ambiguous labels, each event \(e\) may include a \emph{grey zone} interval \(\mathcal G(e)=[g_{\min}(e),g_{\max}(e)]\) where \(g_{\min}(e)\) is the invention date and \(g_{\max}(e)\) is the first recorded use in the White House (i.e., by a sitting president). Any president \(p\) whose term window \([a(p),b(p)]\) overlaps \(\mathcal G(e)\) is \emph{excluded for that event} at data-construction time (i.e., the pair \((p,e)\) is dropped rather than labeled true/false). This prevents false positives/negatives that arise from the time discrepancy. This also helps us remove boundary ambiguity about sporadic or ceremonial access surrounding the installation.

After excluding grey-zone and special-case pairs, we assign the groundtruth feasibility label
\[
y(p,e)\;=\;
\mathbbm{1}\!\Big(\,[a(p),b(p)]\ \cap\ [t_{\min}(e),\,t_{\max}(e)]\ \neq\ \varnothing\Big),
\]
i.e., event \(e\) is \emph{possible} for the president \(p\) iff the presidential term intersects its interval.

For our second variant where we tested the notion of overlapping timelines, each item is an \emph{unordered} group \(S=\{p_1,\dots,p_i\}\) of presidents of size \(i\) (we tested \(N\in\{2,3,4\}\)). Let \(\ell(p)\) and \(d(p)\) denote the birth and death years of president \(p\). Define the group overlap interval to be
\[
I(S)\;=\;\bigcap_{p\in S} [\,\ell(p),\, d(p)\,] 
\;=\; \bigl[\, \max_{p\in S}\ell(p),\; \min_{p\in S}d(p) \,\bigr].
\]
We set the ground-truth label after obtaining birth and death data for each president as
\[
y(S)\;=\;\mathbbm{1}\!\left[\, I(S)\neq\varnothing \,\right]
\;=\;\mathbbm{1}\!\left[\, \max_{p\in S}\ell(p)\;\le\;\min_{p\in S}d(p)\,\right].
\]

Intuitively, \(y(S)=1\) iff the presidents' lifetimes overlap in at least one calendar year, which we obtain by asking \emph{``Were presidents in \(S=\{p_1,\dots,p_i\}\) all alive at the same time''}.

Each trial draws a configurable number (\(N\)) of pairs \((p,e)\) \emph{sampled with replacement} from groundtruth dataset under three batch types (uniformly at random, 100 trails per type):
\begin{enumerate}
  \item \(N / 2\) possible \(+\) \(N / 2\) impossible,
  \item all \(N\) possible,
  \item all \(N\) impossible.
\end{enumerate}
This helps reduce class imbalance and lets us probe evaluation metrics on different categorical mixes.

As before, we utilized GPT-4.1 at \(T{=}0.0\) for determinism. The model receives exactly $N$ statements and must output \texttt{Possible} or \texttt{Not possible} for
each, without explanation or additional text in order for us to make use of regex to capture its boolean responses.

\subsection{Evaluation metrics, deduplication, and aggregation}

We process the results by eliminating duplicate events to ensure each unique president-event pair is only counted once. This is necessary because our sampling method allows the same president-event pair to appear in multiple batches, and counting duplicates would artificially inflate our overall performance metrics. After deduplication, we compute all evaluation metrics on this clean dataset.

Each instance is labeled \(\,y \in \{0,1\}\) as \(\text{Possible}=0\) or \(\text{Not\,possible}=1\).
Predictions \(\hat y\) are obtained from the model’s string outputs after regex pattern capture. Precision/recall treat \(1\) (\text{Not possible}) as the positive class.

Let \(n\) be the number of unique (deduplicated) instances, \( \text{TP}=\sum \mathbbm{1}[y{=}1,\hat y{=}1]\), \( \text{FP}=\sum \mathbbm{1}[y{=}0,\hat y{=}1]\), \( \text{TN}=\sum \mathbbm{1}[y{=}0,\hat y{=}0]\), and \( \text{FN}=\sum \mathbbm{1}[y{=}1,\hat y{=}0]\).
We report:
\[
\text{Accuracy}=\frac{\text{TP}+\text{TN}}{n},\quad
\text{Precision}=\frac{\text{TP}}{\text{TP}+\text{FP}},\quad
\text{Recall}=\frac{\text{TP}}{\text{TP}+\text{FN}},\quad
\text{F1}=\frac{2\,\text{Precision}\cdot\text{Recall}}{\text{Precision}+\text{Recall}}.
\]
We also include the \(2{\times}2\) confusion matrix with rows = ground truth and columns = prediction (\(\begin{psmallmatrix}\text{TN}\ \text{FP}\\ \text{FN}\ \text{TP}\end{psmallmatrix}\)).
When a denominator is zero (e.g., no predicted positives), the corresponding metric is set to \(0\) (\texttt{zero\_division=0} in implementation).

In addition to the overall metrics, we group the deduplicated table by \texttt{batch\_type}
(\textit{half-half}, \textit{all-true}, \textit{all-false}) and compute the same set of statistics per batch type.
For each batch type we also record \((\text{TP},\text{FP},\text{TN},\text{FN})\) and number of president-event pairs \(N\), aggregating over 100 trails per type.

\subsection{Key findings}

\paragraph{Variant 1: Single–boundary feasibility}

\begin{table}[htbp]
\centering
\caption{Variant 1 GPT-4.1 Anachronism Detection Results by Batch Size and Type}
\label{tab:anachronism_results}
\begin{tabular}{l l c c c c}
\toprule
Batch Size & Experiment Type & \multicolumn{4}{c}{Metrics (After Deduplication)} \\
\cline{3-6}
& & Accuracy & Precision & Recall & F1 Score \\
\midrule
\multirow{3}{*}{N = 6}  & 3\_true\_3\_false & 0.998 & 1.000 & 1.000 & 1.000 \\
                        & 6\_false          & 0.995 & N/A   & N/A   & N/A   \\
                        & 6\_true           & 0.995 & 1.000 & 0.974 & 0.987 \\
\midrule
\multirow{3}{*}{N = 10} & 10\_false         & 0.950 & N/A   & N/A   & N/A   \\
                        & 10\_true          & 0.908 & 1.000 & 0.890 & 0.942 \\
                        & 5\_true\_5\_false & 0.932 & 0.933 & 0.933 & 0.933 \\
\midrule
\multirow{3}{*}{N = 20} & 10\_true\_10\_false & 0.992 & 1.000 & 0.989 & 0.994 \\
                        & 20\_false           & 0.982 & N/A   & N/A   & N/A   \\
                        & 20\_true            & 0.982 & 1.000 & 0.983 & 0.991 \\
\bottomrule
\end{tabular}

\par\smallskip
\begin{minipage}{0.92\linewidth}
\footnotesize\emph{Note.} “N/A” indicates cases where precision, recall, and F1 are undefined because there are no positive predictions (all-false batches). These batches can still achieve high accuracy (e.g., $>0.95$) but cannot be evaluated on precision/recall metrics.
\end{minipage}
\end{table}

In single-boundary tests, the anachronism detection results achieved excellent performance across batch sizes, with overall accuracy ranging from 95.0\% to 99.8\%. The results show that the model can reliably distinguish between temporally possible and impossible events with a well-defined first possible dates involving U.S. presidents. We also observe that mixed batches consistently outperform both entirely true and false batches; balanced true/false statements distributions in one prompt are found to be the most reliable evaluation of the model capability. In terms of scalability with increasing batch size (i.e. president-event pair), performance remains consistently high even as batch size increases from 6 to 20. The results demonstrate that the model has effectively learned to detect anachronisms well, especially for events that have no end date and are currently chronologically possible.

Despite the high overall performance, several systematic errors reveal interesting patterns in the model's temporal reasoning.  Table \ref{tab:model_errors} displays both notable false positives (model incorrectly flagged possible events as anachronisms) and false negatives cases (model missed actual anachronisms). For the former, the LLM appears \emph{overly conservative} about early technology adoption, incorrectly flagging events that were chronologically possible, in most cases might be due to temporal proximity. Grant used telephones in San Francisco (1877), but the White House didn't have phones until 1879. This reflects confusion between general telephone technology (available 1877) and White House installation (1879). The photography errors for Taylor and Fillmore suggest the model underestimates how early photography became practical (it was available by the 1840s). Railroads were already operational during Harrison's brief presidency, thus  making it possible for him to travel by railroad then.

For false negatives, the model shows leniency toward temporal boundaries, particularly for recent technologies. John Quincy Adams' presidency (1825-1829) predates practical photography (1839), but the model may be too permissive with "close calls." The generative AI errors for Obama and Bush reveal the model's difficulty with cutting-edge technology timelines, as both served well before generative AI became mainstream (around 2020). 

\begin{table}[htbp]
\centering
\caption{Variant 1 Error Analysis: False Positives vs. False Negatives}
\label{tab:model_errors}
\begin{subtable}[b]{0.48\textwidth}
\centering
\caption{False Positives}
\label{tab:false_positives}
\begin{tabular}{p{0.9\textwidth}}
\toprule
\textbf{Statement} \\
\midrule
Ulysses S. Grant Used the White House telephone \\
Zachary Taylor Appeared in a photograph while president \\
William Henry Harrison Travelled by railroad while president \\
Millard Fillmore Appeared in a photograph while president \\
\bottomrule
\end{tabular}
\end{subtable}
\hfill
\begin{subtable}[b]{0.48\textwidth}
\centering
\caption{False Negatives}
\label{tab:false_negatives}
\begin{tabular}{p{0.9\textwidth}}
\toprule
\textbf{Statement} \\
\midrule
John Quincy Adams Appeared in a photograph while president \\
Barack Obama Used generative AI while president \\
George W. Bush Used generative AI while president \\
\bottomrule
\end{tabular}
\end{subtable}
\end{table}

We also tried prompting the LLM to answer whether a historical figure could have exchanged a letter with a given president. This setup introduced substantial ambiguity: prompts conflated \emph{chronological possibility} (lifespans overlap) with \emph{historical plausibility} (would such an exchange actually occur). For example, in the case of Rutherford B.\ Hayes and Dwight D.\ Eisenhower, GPT-4.1 first answered, ``No---they couldn't have met,'' then immediately noted a narrow overlap window (1890--1893) and concluded that it was ``chronologically possible (their lives overlapped briefly) but logically implausible that they met in person.''

To remove this confound, we move to Variant 2, which asks simply whether all presidents in a group \emph{were alive at the same time}, removing the complexity of evaluating event feasibility and focusing purely on chronological viability.

\paragraph{Variant 2: Multi-timeline overlap}

The presidents overlap experiment asks a more direct question and evaluates the model's ability to determine whether multiple U.S. presidents could have been alive simultaneously. We present the model with groups of $n$ = 2, 3, or 4 presidents and ask: "Were [n presidents] all alive at the same time? Answer Yes or No for each group." For example, given the group "George Washington, John Adams, Thomas Jefferson," the model must determine if there was any point in time when all three presidents were living at the same time. We test 2 batch sizes for each n (N = 6, 10 as before) to see the scaling pattern across different combinatorial complexities. The experiment tests three levels of complexity: 43C2 (903 unique pairs), 43C3 (12,341 unique triplets), and 43C4 (123,410 unique quadruplets), where 43 represents the total number of U.S. presidents and C2-C4 indicates choosing 2, 3, or 4 presidents from this pool. 

\begin{table}[htbp]
\centering
\caption{Variant 2 GPT-4.1 Anachronism Detection Results by Combinatorial Complexity and Batch Type}
\label{tab:presidents_overlap_results}
\begin{minipage}{0.95\linewidth}
\begin{tabular}{lccccc}
\toprule
Combinatorial & Experiment Type & \multicolumn{4}{c}{Metrics (After Deduplication)} \\
\cline{3-6}
Complexity & & Accuracy & Precision & Recall & F1 Score \\
\midrule
\multirow{6}{*}{2 presidents (43C2)} & 3\_true\_3\_false & 0.950 & 0.964 & 0.939 & 0.951 \\
& 6\_false & 0.944 & N/A & N/A & N/A \\
& 6\_true & 0.866 & 1.000 & 0.866 & 0.928 \\
\cline{2-6}
& 10\_false & 0.950 & N/A & N/A & N/A \\
& 10\_true & 0.908 & 1.000 & 0.890 & 0.942 \\
& 5\_true\_5\_false & 0.932 & 0.933 & 0.933 & 0.933 \\
\midrule
\multirow{6}{*}{3 presidents (43C3)} & 3\_true\_3\_false & 0.910 & 0.943 & 0.870 & 0.905 \\
& 6\_false & 0.929 & N/A & N/A & N/A \\
& 6\_true & 0.797 & 1.000 & 0.797 & 0.887 \\
\cline{2-6}
& 10\_false & 0.912 & N/A & N/A & N/A \\
& 10\_true & 0.766 & 1.000 & 0.750 & 0.857 \\
& 5\_true\_5\_false & 0.914 & 0.936 & 0.879 & 0.906 \\
\midrule
\multirow{6}{*}{4 presidents (43C4)} & 3\_true\_3\_false & 0.854 & 0.952 & 0.743 & 0.835 \\
& 6\_false & 0.955 & N/A & N/A & N/A \\
& 6\_true & 0.624 & 1.000 & 0.624 & 0.768 \\
\cline{2-6}
& 10\_false & 0.930 & N/A & N/A & N/A \\
& 10\_true & 0.656 & 1.000 & 0.656 & 0.793 \\
& 5\_true\_5\_false & 0.881 & 0.954 & 0.798 & 0.869 \\
\bottomrule
\end{tabular}

\medskip
\footnotesize \textit{Note.} N/A indicates cases where precision, recall, and F1 are undefined due to no positive predictions (all-false batches). 43C2, 43C3, and 43C4 denote choosing 2, 3, or 4 presidents from a pool of 43.
\end{minipage}
\end{table}

Results in Table \ref{tab:presidents_overlap_results} show a 
performance decline as the number of presidents increases, despite the overall accuracy still being fairly high across all 
set sizes. This supports our hypothesis that the degradation reveals the model's limitations in handling increasingly complex multi-timeline temporal reasoning tasks. The exponential increase in possible 
overlapping combinations of presidents might start to overwhelm the model's chronological ordering ability. The LLM appears to handle pairwise overlapping lifetime relationships well but begins to falter when required to simultaneously evaluate three or four overlapping lifespans, suggesting fundamental limitations in reasoning about overlapping lifespan of multiple presidents rather than simple knowledge gaps. The model also performs 
worse on all-true batches compared to mixed batches across all combinatorial complexities, implying that the model generally can spot temporal impossibility well but becomes less reliable when confirming larger groups where all presidents should theoretically overlap. We do not observe significant bottlenecks or performance degradation when increasing the number of statements or batch size in a prompt from 6 to 10. This suggests that the model's chronological reasoning capabilities are not significantly impacted by processing more statements simultaneously, but by encountering increasing combinatorial complexity (number of presidents being evaluated) within each statement.

\newpage

\section{Conclusion}

Our experiments show that there are fundamental limitations in LLM's understandings of chronology. A core theme emerging from our study is that today’s LLMs have a rudimentary sense of time, and that this understanding is quickly overwhelmed as problem complexity grows. Compared with other problem domains, complexity becomes a challenge for chronological tasks even at relatively small scales.
The consequences are practical: without deliberate temporal reasoning and verification, LLMs cannot realize their potential as tools to aid forecasting, as they remain vulnerable to look-ahead bias. Our findings indicate that there is no prompt-only shortcut to eliminating look-ahead bias: if a model cannot reliably handle basic chronological ordering, leakage will persist. 
But our results also suggest a more promising path through \emph{reasoning modes} that explicitly allocate computation to think internally and verify timelines.

Despite its scope, our study has clear limitations and points to several follow-ups. First, GPT-5's routing to reasoning is itself a challenge. For users who heavily use the chat interface which may or may not auto-route to “thinking” modes, it is unclear whether seemingly simple ordering tasks will be recognized as requiring extra deliberation by ChatGPT 5. Problems will occur if GPT-5 rarely routes to a reasoning mode on its own. Future work should test this explicitly and build an \emph{adaptive} pipeline: run a cheap pass, apply a simple \emph{chronology gate} (“as-of-t” checks, disagreement/uncertainty thresholds), and escalate only when needed (e.g., higher reasoning effort in GPT-5, Extended Thinking in Claude Sonnet). 

Second, our model coverage is still incomplete, despite our attempt at selecting each model which represents a broad coverage of all types available in the market: non-reasoning (GPT-4.1), reasoning (Claude 3.7 Sonnet), reasoning with internal CoT (Claude 3.7 Sonnet with Extended thinking), or even all-in-one model like GPT-5. We tested a subset of GPT-5 settings and did not systematically evaluate open-source reasoning models. A broader sweep across families (e.g., Llama Mixtral-class, and other open “reasoning” variants) should report \emph{cost–accuracy} trade-offs and when escalation actually enhances performance. Third, although our tasks are designed to be domain-agnostic, much of the evaluation relies on historical events; adding domain-specific timelines for specific applications or fields (finance, science, multilingual), as well as cross-domain checks will test generalizability. Finally, beyond single-pass outputs, consensus methods such as LLM-as-judge, self-consistency/majority vote may be helpful in boosting performance. Practical controls should therefore pair time-sliced retrieval prompting or data “fences’’ with explicit deliberate reasoning. Concretely, researchers should: (i) \emph{utilize reasoning modes} for chronology-sensitive tasks (e.g., GPT-5 with higher reasoning effort, Claude Sonnet with extended thinking), and require the model to enumerate and check timelines before answering; (ii) \emph{add a chronology gate} that asks the model to justify that each fact was knowable as of a specific date \(t\), and to instruct models to abstain from making any statements or forecast when uncertainty is high and (iii) \emph{evaluate its decisions} with other reasoning models (LLMs as a judge) or self-consistency/majority vote over multiple runs and models, including auditing with our chronology experiments. 

More broadly during pretraining, building chronologically consistent models will likely require objectives and infrastructure that encode time such as a temporally indexed corpus of data for training LLMs, timeline-aware constraints/regularizers, uncertainty flags/responses for “not knowable as of time t” and robust post-training finetuning tests on feasibility, ordering, and overlap, so that instructions like “answer as if it were 2016’’ are not just arbitrarily prompted but mechanistically respected by the LLM.

\newpage

\appendix

\section{Technical Appendices and Supplementary Material}
\subsection{Evaluation metrics definitions}
\label{app:metrics}
Our baseline experiments evaluate chronological ordering ability on shuffled lists of historical events whose correct order is fully known.  Before presenting the aggregate numbers,
we define the evaluation metrics used throughout this work: Spearman’s \(\rho\), Kendall's \(\tau\), Cayley Distance, and Exact match rate (EMR).

\paragraph{Spearman’s rank correlation coefficient ($\rho$).}
Let $\sigma=(\sigma_1,\dots,\sigma_n)$ be the ground‑truth order and
$\hat\sigma=(\hat\sigma_1,\dots,\hat\sigma_n)$ the predicted order.
Denote by $r_i$ and $\hat r_i$ the ranks of item~$i$ in
$\sigma$ and $\hat\sigma$, respectively.  
\[
\rho \;=\;
1-\frac{6\sum_{i=1}^{n}(r_i-\hat r_i)^2}{n\,(n^{2}-1)}.
\]

We chose Spearman's rank correlation because it gives us a global 
view of ordering quality. 
Whereas Pearson correlation measures linear relationships,
Spearman's measure depends on ranks but not numerical values.
It is sensitive to large positional errors, 
and one misplaced item far from its true slot is heavily penalized. 
It is a well-established statistical measure and is fast to compute.

\paragraph{Kendall's  \(\tau\).}
Let $C$ and $D$ be the numbers of \emph{concordant} (correctly ordered) and
\emph{discordant} (incorrectly ordered) pairs between $\sigma$ and $\hat\sigma$. Then
\[
\tau \;=\;\frac{C-D}{\binom{n}{2}}.
\]

Kendall's  \(\tau\) is robust for small number of events; 
it is less sensitive to single catastrophic errors, as every inversion counts only once.

\paragraph{Cayley distance ($d_{\text{Cayley}}$).}
View $\hat\sigma$ as a permutation
$\pi\in S_n$ that maps the true index of each item to its predicted index.
If $c(\pi)$ is the number of disjoint cycles in the cycle decomposition of
$\pi$, then
\[
d_{\text{Cayley}}(\sigma,\hat\sigma) \;=\; n-c(\pi).
\]

Cayley distance provides a direct physical interpretation of “how many pairwise swaps are wrong?” It grows linearly with the number of events (n) in expectation, giving a scale‑sensitive error that the above two metrics (both bounded in
[-1,1]) cannot provide. It is also strictly zero only when the entire sequence is correct. It is a good complement to the exact‑match rate.

\paragraph{Normalized Cayley distance.}
To make distance scores comparable across lists of different lengths, the Cayley distance \(d_{\text{Cayley}}\)—the minimum number of adjacent transpositions required to transform the model’s permutation into the ground
truth—we scale it to get
\[
\text{Cayley}_{\mathrm{norm}}
  = \frac{d_{\text{Cayley}}}{n-1}, \qquad 0\le
    \text{Cayley}_{\mathrm{norm}}\le 1.
\]
This normalization enables meaningful cross‑\(n\) comparisons while preserving the interpretability of~\(d_{\text{Cayley}}\).

\paragraph{Exact‑match rate (EMR).}
Given $T$ independent trials, let
\(
\sigma^{(t)},\hat\sigma^{(t)}\in S_{n_t}
\)
denote the ground‑truth and predicted permutations in trial $t$.
Define the indicator
\(
\mathbb{I}\!\bigl[\hat\sigma^{(t)}=\sigma^{(t)}\bigr]
=
\mathbb{I}\!\bigl[d_{\text{Cayley}}\!\bigl(\sigma^{(t)},\hat\sigma^{(t)}\bigr)=0\bigr].
\)
The exact‑match rate is the empirical mean of this indicator:
\[
\text{EMR}
\;=\;
\frac{1}{T}\sum_{t=1}^{T}
\mathbb{I}\!\bigl[\hat\sigma^{(t)}=\sigma^{(t)}\bigr].
\]
It equals the proportion of trials in which the predicted list is
\emph{identical} to the ground truth and, equivalently, the Cayley distance
is zero.

\subsubsection{Position-wise error metric}
\label{app:metrics:positionwise}

\paragraph{Mean Absolute Rank Difference (MARD)}
Let trials be indexed by \(t\) with list length \(n_t\). For an item \(e\) in trial \(t\), denote its true and predicted ranks by
\(r_{\text{true}}^{(t)}(e), \hat r^{(t)}(e)\in\{1,\dots,n_t\}\).
Fix a list length \(n\) and a ground-truth position \(k\in\{1,\dots,n\}\), and define
\[
\mathcal I_{n,k} \;=\; \bigl\{(t,e):\, n_t=n,\; r_{\text{true}}^{(t)}(e)=k \bigr\}.
\]
The \emph{mean absolute rank difference (MARD)} at position \(k\) is
\[
\mathrm{MARD}_{n}(k)
\;=\;
\frac{1}{\lvert \mathcal I_{n,k}\rvert}
\sum_{(t,e)\in \mathcal I_{n,k}}
\bigl|\, \hat r^{(t)}(e) - k \,\bigr|.
\]
This measure takes values between 0 and $n-1$, with lower values indicating smaller errors.

\paragraph{Per-position accuracy}
\label{app:metrics:perpos-acc}

Let criteria \(P\) specify the target set \(G_P=\{i_1,\dots,i_n\}\) (e.g., \textsc{ABC-First-Names}, \textsc{OhioOrVirginia}), with
ground-truth order \(\pi^{\text{true}}=(i_1,\dots,i_n)\).
For trial \(t\), let the model’s prediction be \(\pi^{(t)}=(i^{(t)}_1,\dots,i^{(t)}_{m_t})\).
We condition on the subset of trials with no length mismatch:
\[
\mathcal T_{n,P}\;=\;\{\,t:\; m_t=n\,\}.
\]

For rank \(r\in\{1,\dots,n\}\), define the exact-occupant accuracy
\[
A_{n,P}(r)\;=\;\frac{1}{|\mathcal T_{n,P}|}\sum_{t\in\mathcal T_{n,P}}
\mathbf 1\!\left[\, i^{(t)}_{r}=i_{r}\,\right].
\]
This measure takes values between 0 and 1, with higher values indicating greater accuracy.

\newpage

\subsection{Datasets}
\label{app:data}

This appendix describes the datasets used in our ordering experiments.

\subsubsection{20th Century Historical Events}

We use a dataset of 20th century historical events extracted from the \href{https://en.wikipedia.org/wiki/Timeline_of_the_20th_century}{Wikipedia Timeline of the 20th Century}. The events were scraped and processed into a structured CSV format, which can be accessed \href{https://drive.google.com/file/d/1-kWBPQlvFGEQo1SicFjH_RgzsUmwx7Al/view?usp=sharing}{here}. This dataset contains major historical events spanning from 1901 to 2000, covering political, social, technological, and cultural milestones that shaped the 20th century.

\subsubsection{Wide Time-Scale Historical Events}

Table~\ref{tab:wide_events} presents our wide time-scale historical events dataset used in the wide-gap variant for basic sorting on events experiment.

\begin{table}[htbp]
\centering
\caption{Wide Time-Scale Historical Events Dataset}
\label{tab:wide_events}
\vspace{\baselineskip}
\begin{tabular}{cl}
\toprule
\textbf{Year} & \textbf{Event} \\
\midrule
43 & Southern Britain annexed by Rome \\
161 & Death of Antoninus Pius \\
380 & Christianity becomes official religion of Roman Empire \\
476 & Fall of Western Roman Empire \\
581 & China is unified by the Sui Dynasty \\
697 & Venice becomes independent from Eastern Roman Empire \\
762 & Baghdad founded; becomes center of learning during Islamic Golden Age \\
808 & Gunpowder discovered in China \\
927 & Kingdom of England established by Æthelstan \\
1096 & Oxford University begins functioning \\
1271 & Yuan Dynasty established in China by Kublai Khan \\
1452 & Birth of Leonardo da Vinci; Renaissance begins \\
1511 & Spanish Conquest of America begins \\
1643 & Birth of Isaac Newton; start of the Enlightenment \\
1707 & Union of England and Scotland \\
1803 & Napoleon sells Louisiana Territory to USA \\
1826 & Photography invented by Joseph Nicéphore (France) \\
1905 & Einstein creates E=mc²; basis for atomic energy \\
1919 & Treaty of Versailles \\
1949 & Creation of NATO \\
1971 & Pentagon Papers leaked; leads to US withdrawal from Vietnam \\
1993 & European Union founded \\
2000 & Israeli troops withdraw from southern Lebanon \\
2001 & 9/11 attacks in the US \\
2005 & YouTube is founded \\
2012 & Curiosity rover takes selfie and finds ancient water streambed on Mars \\
2016 & UK votes to leave the EU; beginning of Brexit \\
2020 & Global COVID-19 pandemic begins \\
2022 & Human population reaches 8 billion \\
2025 & Pope Francis dies at 88 \\
\bottomrule
\end{tabular}
\vspace{\baselineskip}
\end{table}

\subsubsection{U.S. Presidents List}

Table~\ref{tab:presidents_sample} shows all 43 US presidents used in the experiment.

\begin{table}[ht]
  \centering
  \caption{Presidential Birth Information and Terms}
  \label{tab:presidents_sample}
  \setlength{\tabcolsep}{4pt} 

  \begin{tabularx}{\linewidth}{@{}Y l c Y r r@{}}
    \toprule
    \textbf{Name} & \textbf{Birthdate} & \textbf{\shortstack{Birth\\Day of Week}} &
    \textbf{Birth State} & \textbf{Start} & \textbf{End} \\
    \midrule
    George Washington & 1732-02-22 & Friday & Virginia & 1789 & 1797 \\
    John Adams & 1735-10-30 & Sunday & Massachusetts & 1797 & 1801 \\
    Thomas Jefferson & 1743-04-13 & Saturday & Virginia & 1801 & 1809 \\
    James Madison & 1751-03-16 & Tuesday & Virginia & 1809 & 1817 \\
    James Monroe & 1758-04-28 & Friday & Virginia & 1817 & 1825 \\
    John Quincy Adams & 1767-07-11 & Saturday & Massachusetts & 1825 & 1829 \\
    Andrew Jackson & 1767-03-15 & Sunday & South Carolina & 1829 & 1837 \\
    Martin Van Buren & 1782-12-05 & Thursday & New York & 1837 & 1841 \\
    William Henry Harrison & 1773-02-09 & Tuesday & Virginia & 1841 & 1841 \\
    John Tyler & 1790-03-29 & Monday & Virginia & 1841 & 1845 \\
    James K. Polk & 1795-11-02 & Monday & North Carolina & 1845 & 1849 \\
    Zachary Taylor & 1784-11-24 & Wednesday & Virginia & 1849 & 1850 \\
    Millard Fillmore & 1800-01-07 & Tuesday & New York & 1850 & 1853 \\
    Franklin Pierce & 1804-11-23 & Friday & New Hampshire & 1853 & 1857 \\
    James Buchanan & 1791-04-23 & Saturday & Pennsylvania & 1857 & 1861 \\
    Abraham Lincoln & 1809-02-12 & Sunday & Kentucky & 1861 & 1865 \\
    Andrew Johnson & 1808-12-29 & Thursday & North Carolina & 1865 & 1869 \\
    Ulysses S. Grant & 1822-04-27 & Saturday & Ohio & 1869 & 1877 \\
    Rutherford B. Hayes & 1822-10-04 & Friday & Ohio & 1877 & 1881 \\
    James A. Garfield & 1831-11-19 & Saturday & Ohio & 1881 & 1881 \\
    Chester A. Arthur & 1829-10-05 & Monday & Vermont & 1881 & 1885 \\
    Benjamin Harrison & 1833-08-20 & Tuesday & Ohio & 1889 & 1893 \\
    William McKinley & 1843-01-29 & Sunday & Ohio & 1897 & 1901 \\
    Theodore Roosevelt & 1858-10-27 & Wednesday & New York & 1901 & 1909 \\
    William Howard Taft & 1857-09-15 & Tuesday & Ohio & 1909 & 1913 \\
    Woodrow Wilson & 1856-12-28 & Sunday & Virginia & 1913 & 1921 \\
    Warren G. Harding & 1865-11-02 & Thursday & Ohio & 1921 & 1923 \\
    Calvin Coolidge & 1872-07-04 & Thursday & Vermont & 1923 & 1929 \\
    Herbert Hoover & 1874-08-10 & Monday & Iowa & 1929 & 1933 \\
    Franklin D. Roosevelt & 1882-01-30 & Monday & New York & 1933 & 1945 \\
    Harry S. Truman & 1884-05-08 & Thursday & Missouri & 1945 & 1953 \\
    Dwight D. Eisenhower & 1890-10-14 & Tuesday & Texas & 1953 & 1961 \\
    John F. Kennedy & 1917-05-29 & Tuesday & Massachusetts & 1961 & 1963 \\
    Lyndon B. Johnson & 1908-08-27 & Thursday & Texas & 1963 & 1969 \\
    Richard Nixon & 1913-01-09 & Thursday & California & 1969 & 1974 \\
    Gerald Ford & 1913-07-14 & Monday & Nebraska & 1974 & 1977 \\
    Jimmy Carter & 1924-10-01 & Wednesday & Georgia & 1977 & 1981 \\
    Ronald Reagan & 1911-02-06 & Monday & Illinois & 1981 & 1989 \\
    George H. W. Bush & 1924-06-12 & Thursday & Massachusetts & 1989 & 1993 \\
    Bill Clinton & 1946-08-19 & Monday & Arkansas & 1993 & 2001 \\
    George W. Bush & 1946-07-06 & Saturday & Connecticut & 2001 & 2009 \\
    Barack Obama & 1961-08-04 & Friday & Hawaii & 2009 & 2017 \\
    Joe Biden & 1942-11-20 & Friday & Pennsylvania & 2021 & 2025 \\
    \bottomrule
  \end{tabularx}
\end{table}

\newpage

Table~\ref{tab:presidents_schema} describes the schema of the above US Presidents dataset.

\begin{table}[htbp]
\centering
\caption{US Presidents Dataset Schema}
\label{tab:presidents_schema}
\vspace{\baselineskip}
\begin{tabular}{lll}
\toprule
\textbf{Column Name} & \textbf{Data Type} & \textbf{Description} \\
\midrule
Name & String & Full name of the US president \\
Birthdate & Date (YYYY-MM-DD) & Birth date in ISO format \\
Birth Day Of Week & String & Day of the week when president was born \\
Birth State & String & State where the president was born \\
Start & Integer & Year when presidency began \\
End & Integer & Year when presidency ended \\
\bottomrule
\end{tabular}
\vspace{\baselineskip}
\end{table}

\subsection{LLM Prompt Templates}
\label{app:prompts}

All runs are closed-book (no external tools/browsing). Prompts begin with a reset instruction to ignore prior context, use only internal pre-trained knowledge, and require an exact output format. We show canonical templates: square brackets \([\cdot]\) mark fields filled programmatically (e.g., sampled items), while curly braces \(\{\cdot\}\) mark user-customizable options (e.g., criteria names, batch size).

\subsubsection{Knowledge-Verification}
\label{app:knowledgeverification}
\begin{lstlisting}[style=prompt]
In what year did this historical event occur? Respond with ONLY the year as a number: [event] (Guidance: If a range exists, answer with the first year. If unknown, output UNKNOWN.)
\end{lstlisting}
\noindent
Used to confirm that the LLM knows the events it is asked to order. Items accepted if and only if the model returns the exact 4-digit canonical year; otherwise the item is excluded.

\subsubsection{Basic Sorting — Historical Events}
\begin{lstlisting}[style=prompt]
You are an expert historian specializing in accurate chronological sequencing. Arrange the events below in strict chronological order (earliest to latest).

[event lists]

TASK RULES
1) If an event spans a range (e.g., "Feb 9–19" or "July–August"), order by the FIRST date/month.
2) If an event lacks day or month, reason at the most precise available granularity.
3) If two events fall on the same exact date, break ties ALPHABETICALLY by description.

Return ONLY the reordered list—one event per line, no commentary.

Example
Input:
- Moon Landing
- World War II ends
- Fall of the Berlin Wall

Output:
World War II ends
Moon Landing
Fall of the Berlin Wall

Now reorder these events chronologically:

[event lists]
\end{lstlisting}

\newpage

\subsubsection{Basic Sorting — U.S. Presidents}
\begin{lstlisting}[style=prompt]
You are an expert historian specializing in accurate chronological sequencing. Arrange the presidents below in strict chronological order by the START YEAR of their presidency. If two presidents somehow share the same start date, break ties alphabetically.

[president lists]

Return ONLY the reordered list—one president per line, no commentary.

Example
Input:
- Abraham Lincoln
- John F. Kennedy
- George Washington

Output:
George Washington
Abraham Lincoln
John F. Kennedy

Now reorder these presidents chronologically:

[president lists]
\end{lstlisting}

\subsubsection{Conditional Sorting — Self-Filtering Task}
\begin{lstlisting}[style=prompt]
You are an expert historian specializing in accurate chronological sequencing. You are given a list of US presidents in random order:

[presidents list]

Your task:
1. Filter for presidents who were {criterion_desc}.
2. Then, order the filtered names chronologically (by when they served as president).

Return ONLY the names of presidents who were {criterion_desc}.

Output format:
- One president per line
- No numbering
- No commentary or explanations

Example:
George Washington
Thomas Jefferson
Ulysses S. Grant
...

Do not include any other presidents. Do not include any explanations or additional text.
\end{lstlisting}

\newpage

\subsubsection{Conditional Sorting — Given-Names Task}
\begin{lstlisting}[style=prompt]
You are an expert historian specializing in accurate chronological sequencing. You are given a list of presidents in random order:

[presidents list]

Your task:
Order these presidents chronologically (by when they served as president).

Return ONLY the names of these presidents, ordered chronologically.

Output format:
- One president per line
- No numbering
- No commentary or explanations

Example:
George Washington
Thomas Jefferson
Ulysses S. Grant
...

Do not include any other presidents. Do not include any explanations or additional text
\end{lstlisting}

\subsubsection{Anachronism Detection: Variant 1}
\begin{lstlisting}[style=prompt]
You are an expert historian who specializes in accurate chronological sequencing. Below are {BATCH_SIZE} statements about U.S. presidents and activities.

For each, respond with 'Possible' if the activity could have occurred during their presidency, or 'Not possible' if not. 

Give NO explanation.Output exactly one line per event, in the  format: [event]: Possible or [event]: Not possible. Here, [event] means the event text, not literal brackets.

Do not number the lines. Do not add any extra commentary or formatting. Respond to only the events listed below.

Example output:
Abraham Lincoln travelled by railroad while president: Possible
George Washington joined a Zoom call while president: Not possible
Franklin D. Roosevelt hosted a radio 'fireside chat' as president: Possible
John Adams used generative AI while president: Not possible
Barack Obama posted on a social media platform while president: Possible
James K. Polk appeared in a photograph while president: Possible

Now, for the following events, respond in the same format:

[presidents-events pairs]

\end{lstlisting}
\newpage

\subsubsection{Anachronism Detection: Variant 2}
\begin{lstlisting}[style=prompt]
You are an expert historian who specializes in accurate chronological sequencing. Below are [BATCH_SIZE] questions about whether [PRES_COUNT] U.S. presidents were all alive at the same time. For each question, respond with 'Yes' if all [PRES_COUNT] presidents were alive at the same time (their lifetimes overlapped), or 'No' if not.

ONLY consider chronological plausibility (whether their lifetimes overlapped), NOT logical plausibility.
Give NO explanation.

Output exactly one line per question, in the format:
[question]: Yes or [question]: No
Here, [question] means the question text, not literal brackets.

Do not number the lines. Do not add any extra commentary or formatting. Respond only to the questions listed below.

Example output:
Were George Washington and John Adams all alive at the same time: Yes
Were George Washington and Barack Obama all alive at the same time: No
Were Thomas Jefferson, James Madison, and James Monroe all alive at the same time: Yes
Were Abraham Lincoln and John F. Kennedy all alive at the same time: No
Were Franklin D. Roosevelt, Harry S. Truman, and Dwight D. Eisenhower all alive at the same time: Yes
Were Dwight D. Eisenhower and Joe Biden all alive at the same time: Yes

Now, for the following questions, respond in the same format:
[QUESTION_1]
[QUESTION_2]
[...]
\end{lstlisting}

\subsection{Random-Permutation Baseline}
\label{app:rand-baseline}
For each \(n\in\{2,5,10,20,50,100\}\) we generated \(1\,000\) uniformly random orderings and computed Spearman’s~\(\rho\), Kendall’s \(\tau\), and \(\text{Cayley}_{\mathrm{norm}}\) for each.  GPT‑4.1’s score was then expressed as an empirical percentile within this distribution (e.g.\ ``97th percentile against chance’’). Table \ref{tab:rand‑baseline} reports the percentile position of GPT‑4.1’s mean score within the corresponding random distribution; higher values indicate a larger margin over chance.

\begin{table}[htbp]
  \centering
  \footnotesize
  \caption{GPT‑4.1 percentile against \(1000\) random permutations
           (higher\,=\,better).}
  \label{tab:rand‑baseline}
  \begin{tabular}{@{}c cccc@{}}
    \toprule
    \(n\) & Spearman's \(\rho\) & Kendall's \(\tau\) &
          Cayley\(_{\text{norm}}\) & Exact match \\
    \midrule
     2  & 74.2\% & 74.2\% & 74.1\% & 74.2\% \\
     5  & 98.3\% & 98.2\% & 96.1\% & 72.1\% \\
    10  & 100.0\%&100.0\% & 99.0\% & 55.0\% \\
    20  & 100.0\%&100.0\% &100.0\% & 50.0\% \\
    50  & 100.0\%&100.0\% & 97.2\% & 50.0\% \\
   100  & 100.0\%&100.0\% &100.0\% & 50.1\% \\
    \bottomrule
  \end{tabular}
\end{table}

\begin{figure}[htbp]
  \centering
  \includegraphics[width=\linewidth]{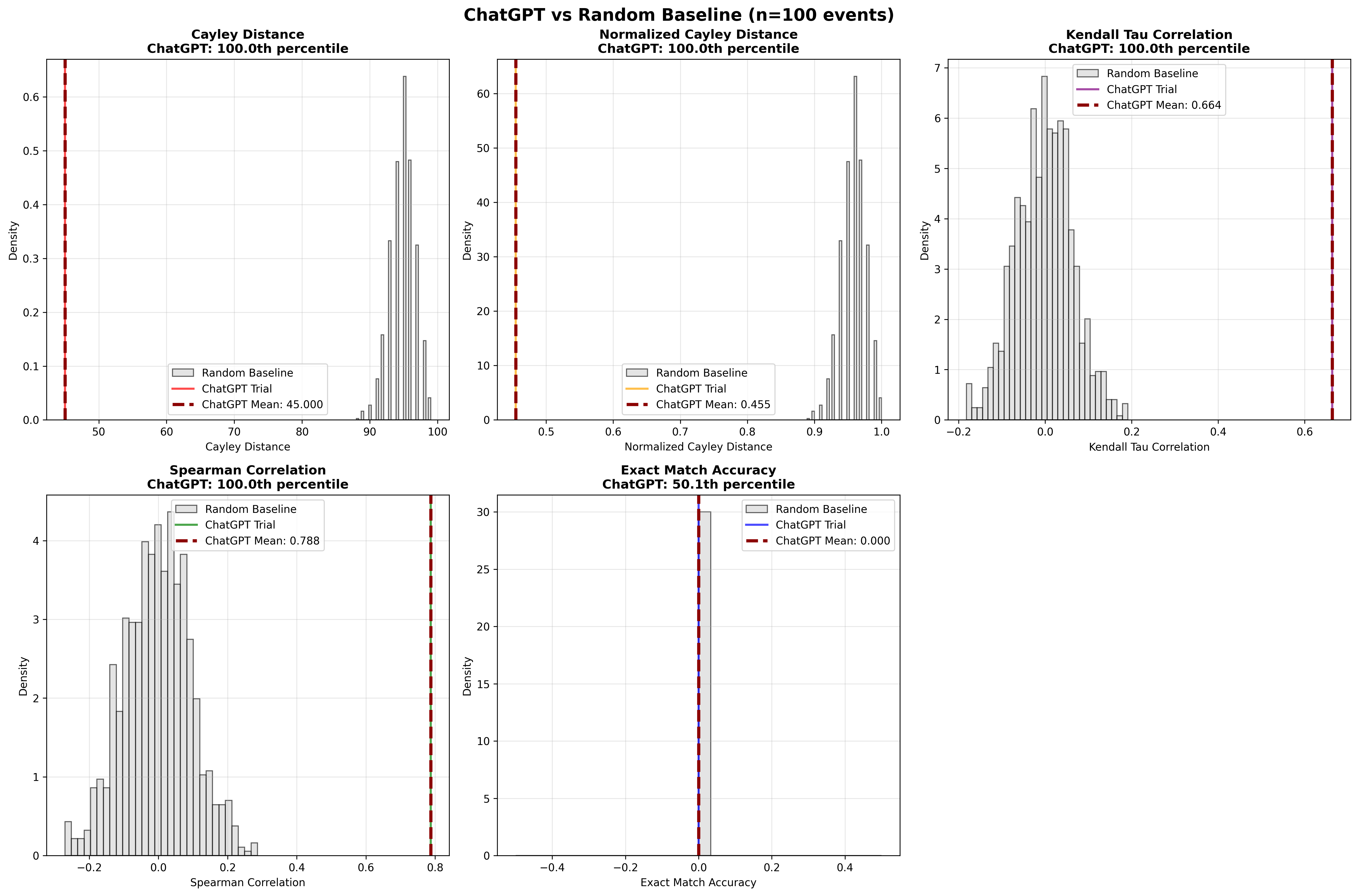}
  \caption{Example visualization of GPT‑4.1 performance versus a random‑permutation baseline for \(n=100\) events. Histograms show the random distribution for each metric (grey bars); the red dashed line marks GPT‑4.1’s mean score and the title of each panel states its percentile rank.}
  \label{fig:n100‑baseline}
\end{figure}

\newpage

\subsection{Basic Sorting Wide Time-Gap Variant}
\label{app:timescale}

\paragraph{Experimental design}
We construct a corpus of \(30\) single–year historical items spanning Years \(1\)–\(2025\), as presented in Table \ref{tab:timescale-dataset-samples}. The goal is broad temporal coverage rather than exhaustiveness: we heuristically aimed to include events from (nearly) every century and a mix of political, scientific, and cultural milestones, but did not enforce a rigid quota per era.  Each item has a canonical year (no multi‑year durations), which enables controlled sampling at prescribed minimum separations \(\Delta\) (e.g., \(50/100/200\) years) in the timescale experiments.

\begin{table}[htbp]
  \centering
  \footnotesize
  \caption{Illustrative samples from the curated dataset (earliest “head” examples at left; latest “tail” examples at right).}
  \label{tab:timescale-dataset-samples}
  \begin{minipage}{0.48\linewidth}
    \centering
    \subcaption{Head (earliest examples)}
    \vspace{0.25em}
    \begin{tabular}{@{}r p{0.70\linewidth}@{}}
      \toprule
      \textbf{Year} & \textbf{Event} \\
      \midrule
       43  & Southern Britain annexed by Rome \\
      161  & Death of Antoninus Pius \\
      380  & Christianity becomes official religion of the Roman Empire \\
      476  & Fall of Western Roman Empire \\
      581  & China unified by the Sui Dynasty \\
      697  & Venice becomes independent from the Eastern Roman Empire \\
      762  & Baghdad founded; later a center of learning in the Islamic Golden Age \\
      808  & Gunpowder discovered in China \\
      927  & Kingdom of England established by \AE{}thelstan \\
     1096  & Oxford University begins functioning \\
     \bottomrule
    \end{tabular}
  \end{minipage}\hfill
  \begin{minipage}{0.48\linewidth}
    \centering
    \subcaption{Tail (latest examples)}
    \vspace{0.25em}
    \begin{tabular}{@{}r p{0.70\linewidth}@{}}
      \toprule
      \textbf{Year} & \textbf{Event} \\
      \midrule
     1993 & European Union founded \\
     2000 & Israeli troops withdraw from southern Lebanon \\
     2001 & 9/11 attacks in the United States \\
     2005 & YouTube is founded \\
     2012 & Curiosity rover finds ancient streambed on Mars \\
     2016 & U.K.\ votes to leave the EU (Brexit begins) \\
     2020 & Global COVID‑19 pandemic begins \\
     2022 & Human population reaches 8~billion \\
     2025 & Pope Francis dies at age 88 \\
     \bottomrule
    \end{tabular}
  \end{minipage}
\end{table}

We aim to roughly simulate \emph{temporal separation} rather than enforce rigid per-century quotas, which could oversample obscure items and introduce knowledge confounds. For our purpose—\emph{relative} comparisons between “wide‑gap’’ and “narrow‑gap’’ regimes—this soft heuristic coverage is sufficient.

\paragraph{Event selection formalization}
We outline the mathematical framework we follow when curating 30 events here.

Let \(\mathcal{U}=\{(y_k,e_k)\}_{k=1}^{30}\) be the curated universe, where \(y_k\in\mathbb{Z}\) is the (single) year of event \(e_k\).
For a trial of size \(n\), we select a subset \(S=\{(y_i,e_i)\}_{i=1}^{n}\) by heuristic sampling that favors broad temporal spread (e.g., drawing from different centuries when feasible).  Define the empirical minimum gap and the fraction of wide pairs:
\[
\widehat{\Delta}_{\min}(S) \;=\; \min_{i<j}\,\lvert y_i-y_j\rvert,
\qquad
\widehat{\pi}_{\Delta}(S) \;=\; \frac{1}{\binom{n}{2}}\sum_{i<j}\mathbf{1}\!\left\{\lvert y_i-y_j\rvert \ge \Delta_{\text{tgt}}\right\}.
\]

Here \(\Delta_{\text{tgt}}\in\{50,100,200\}\) denotes a \emph{target} timescale.  In a fully constrained design one would enforce \(\widehat{\Delta}_{\min}(S)\ge \Delta_{\text{tgt}}\) for every trial.  Instead, we \emph{simulate} the wide‑timescale condition by aiming for
\[
\widehat{\pi}_{\Delta}(S)\;\gtrsim\;p_0 \quad\text{with}\quad p_0\in[0.8,0.95],
\]
i.e., the vast majority of pairs exceed \(\Delta_{\text{tgt}}\), while allowing occasional closer pairs when historical events sampled are obscure.  This approach is sufficient for testing the hypothesis that larger temporal separations ease chronological ordering for LLMs.

Just like other experiments, we first apply knowledge verification (year query at \(T{=}0\)) to obtain the known set
\(\mathcal{K}=\{e_k\in\mathcal{U}:\hat y_k=y_k\}\), then keep one event per year to remove duplicates (final \(|\mathcal{K}|=24\)). For each \(n\in\{2,5,10,15,20,24\}\), we run 20 trials: sample \(n\) events without replacement from \(\mathcal{K}\), permute, and prompt the LLM to order. Prompts and metrics follow the basic-sorting setup (Appendix~\ref{app:prompts}, \ref{app:metrics}).

\newpage

\subsection{Basic Sorting \texorpdfstring{U.\,S.\ Presidents}{US Presidents}}

\subsubsection{Experimental design}
\label{app:pres-method}

To analyze sample‑size effects at fine granularity we generated
\(\,N_{\text{trials}}=20\) independent trials for each of the following number of presidents (list sizes):\[ n \in \{2, 5, 10, 15, 20, 25, 30, 35, 40, 43\}\] yielding a total of \(200\) trials.  For every trial we performed the following steps:

\begin{enumerate}[nosep,leftmargin=1.5em]
  \item \textbf{Sampling:}  Draw \(n\) presidents without replacement from the 43‑person pool.
  \item \textbf{Shuffling:}  Permute the sampled subset uniformly at random.
  \item \textbf{Prompting:}  Present the shuffled list to GPT-4.1 with a standard instruction: \emph{``Order these U.S.\ presidents in the correct chronological order of when they served as presidents.''}
  \item \textbf{Evaluation:}  Score the model’s output with five metrics: exact‑match rate, Spearman’s~\(\rho\), Kendall’s~\(\tau\), Cayley distance, and \(\text{Cayley}_{\mathrm{norm}}\)
\end{enumerate}

All prompts are issued with temperature  \(0.0\) and a fixed random seed (\texttt{42}) to ensure reproducibility.  Events are described only by presidential names—no years, inaugurations, or other temporal cues are included—so the model must rely on its internal chronology rather than
surface hints. Figure \ref{fig:president-workflow} represents the complete workflow for the U.S. Presidents Chronological ordering task.

\tikzset{
  mybox/.style      = {rectangle, rounded corners, draw=black!60,
                       fill=gray!10, font=\footnotesize, align=center,
                       inner sep=4pt},
  mydecision/.style = {diamond, aspect=2, draw=black!60,
                       fill=gray!10, font=\footnotesize, align=center},
  myarrow/.style    = {-{Stealth[length=2mm,width=2mm]}, thick},
}

\begin{figure}[htbp]
  \centering
  \resizebox{0.50\columnwidth}{!}{%
    \begin{tikzpicture}[node distance = 8mm and 14mm]

      \node[mybox] (pool)
        {43\,unique U.S.\ presidents\\(Washington → Biden)};

      \node[mybox, below=of pool] (sample)
        {Sample $n\in\{2,5,10,20,25,30,35,40,43\}$\\without replacement};

      \node[mybox, below=of sample] (shuffle)
        {Uniform shuffle};

      \node[mybox, below=of shuffle] (prompt)
        {Prompt at $T{=}0$:\\\small ``Order these presidents chronologically.''};

      \node[mybox, below=of prompt] (model)
        {GPT‑4.1 output\\(predicted order)};

      \node[mybox, below=of model] (eval)
        {Evaluate:\\ exact‑match,\,$\rho$,\,$\tau$, Cayley,\,$\text{Cayley}_{\text{norm}}$};

      \draw[myarrow] (pool)   -- (sample);
      \draw[myarrow] (sample) -- (shuffle);
      \draw[myarrow] (shuffle) -- (prompt);
      \draw[myarrow] (prompt) -- (model);
      \draw[myarrow] (model)  -- (eval);

    \end{tikzpicture}%
  }
  \caption{Workflow for the U.S.\ Presidents chronology experiment.}
  \label{fig:president-workflow}
\end{figure}
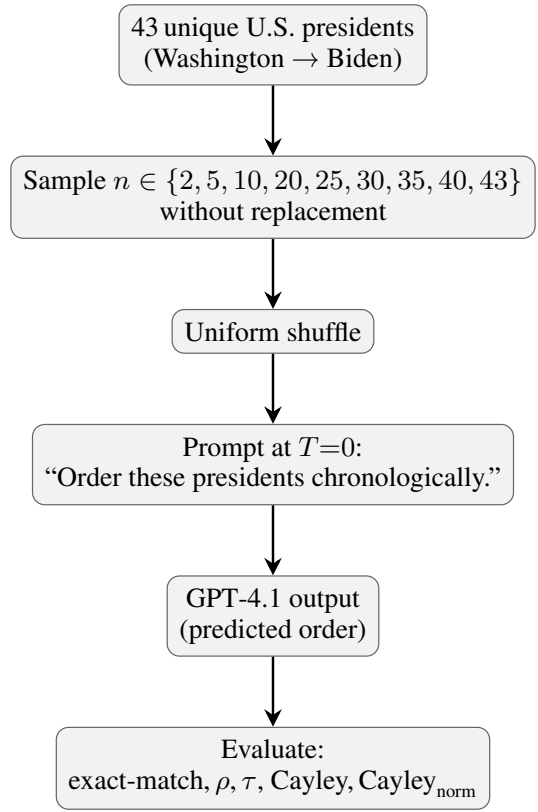

\newpage

\subsubsection{Adapted evaluation pipeline for hallucinated and missing names}
\label{app:adapted-eval}

Let \(n_{\text{prompt}}\) denote the number of presidents that appear in a given prompt, and let
\(\{1,2,\dots,n_{\text{prompt}}\}\) index those names in their ground‑truth order.  

Because the cleaning step deletes any row with a missing or hallucinated
prediction, the rank‑based statistics are computed on a \emph{reduced} set
of indices.  
Let
\[
   \mathcal P \;=\; \{1,2,\dots,n_{\text{prompt}}\}
   \quad\text{be the \textbf{full} index set of presidents in the prompt,}
\]
and let  
\[
   \mathcal V \subseteq \mathcal P
\]
denote the indices that survive the filter.  Writing
\(m = |\mathcal V|\le n_{\text{prompt}}\), the reported correlations are
\[
\rho^{\star} =
\rho\bigl((r_{\text{true},i})_{i\in\mathcal V},
          (r_{\text{pred},i})_{i\in\mathcal V}\bigr),\qquad
\tau^{\star} =
\tau\bigl((r_{\text{true},i})_{i\in\mathcal V},
          (r_{\text{pred},i})_{i\in\mathcal V}\bigr).
\]

Thus a high value of \(\rho^{\star}\) or \(\tau^{\star}\) testifies only
that the \emph{retained} \(m\) presidents are well ordered; it does not
imply that the model produced a faithful chronology for the full
length \(n_{\text{prompt}}\). Correlations should therefore be
read alongside the \textsc{miss} and \textsc{extra} counters as well as the
strict metrics—exact‑match and Cayley distance—that drop to zero whenever
any omission or hallucination occurs.

\newpage
\subsection{Basic Sorting Results Tables \& Figures}
\label{app:basic-sorting}

\subsubsection{GPT-4.1's performance on 20th Century Historical Timeline}
\begin{figure}[htbp]
  \centering
  \includegraphics[width=\linewidth]{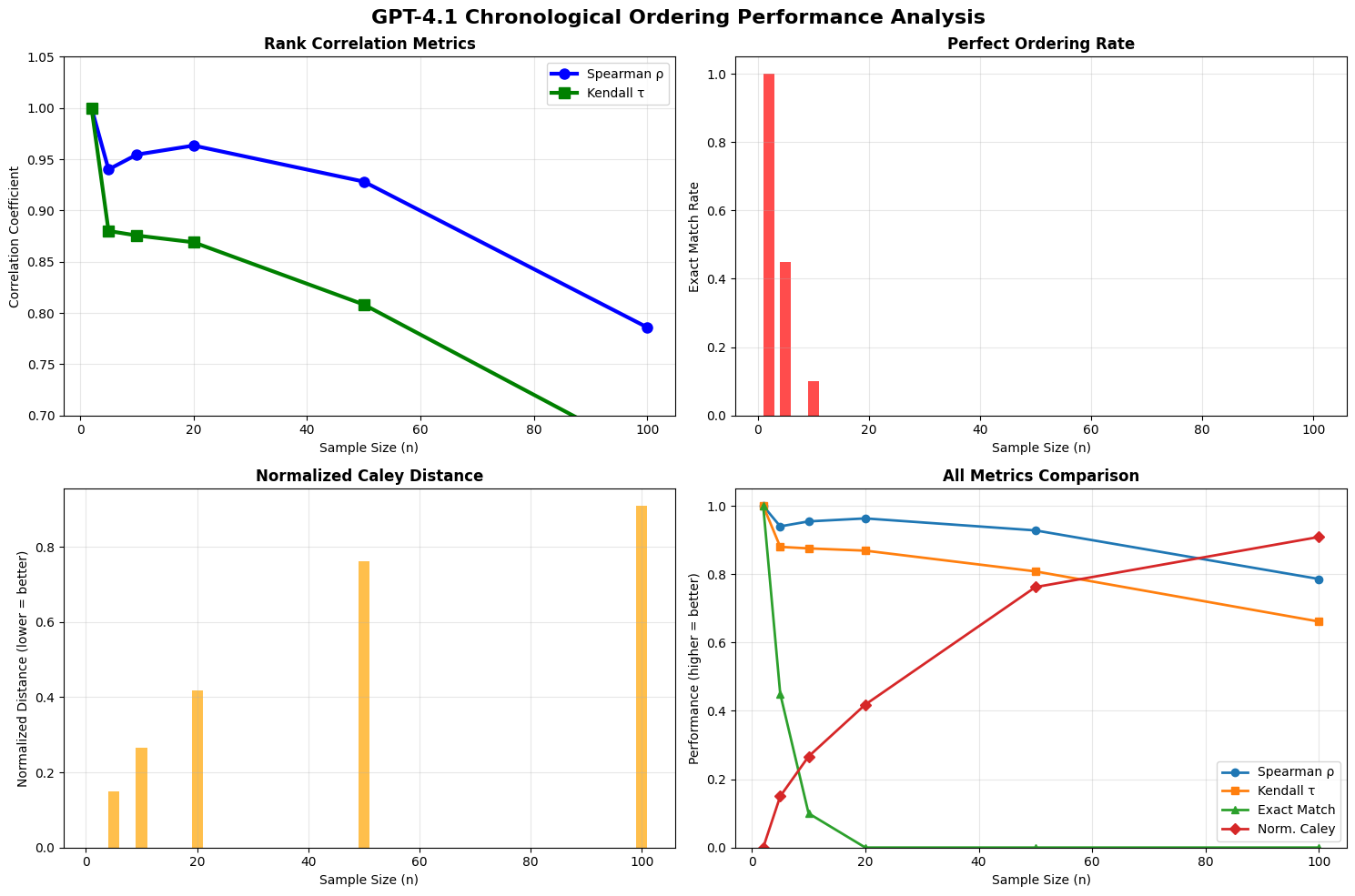}
  \caption{GPT‑4.1 chronological‑ordering performance across list sizes. Top‑left: Spearman’s \(\rho\) and Kendall’s \(\tau\); top‑right: exact‑match rate; bottom‑left: normalized Cayley distance; bottom‑right: all metrics on a common scale. Results are averaged over 20 trials per \(n\).}
  \label{fig:perf‑panels}
\end{figure}

\begin{figure}[htbp]
  \centering
  \includegraphics[width=0.90\linewidth]{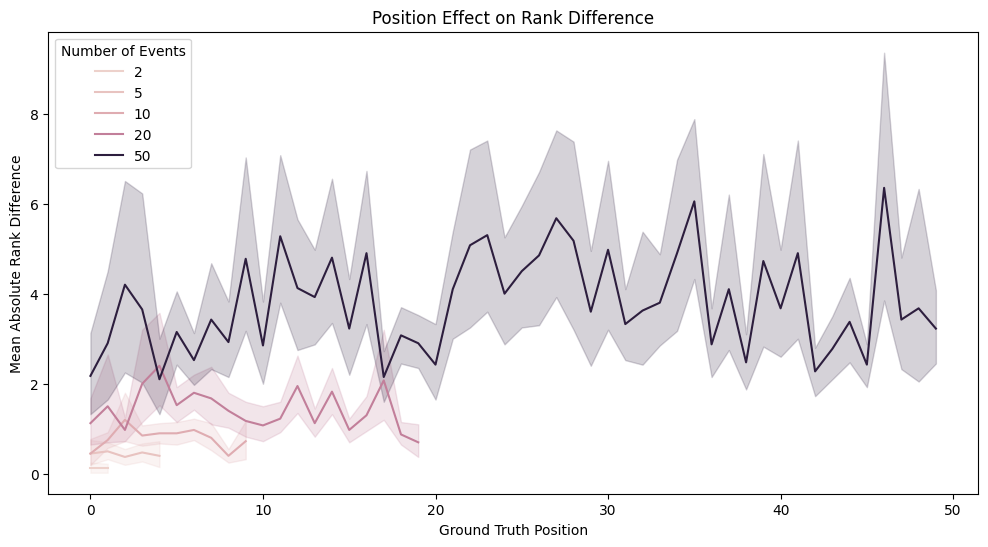}
  \caption{Mean absolute rank difference - MARD (solid line) and inter‑trial variability at each position (shaded band, one standard deviation) as a function of an event’s ground‑truth position in the list.}
  \label{fig:pos‑effect}
\end{figure}

\newpage
\subsubsection{GPT-4.1's performance on Wide Time-gap}

\begin{figure}[htbp]
  \centering
  \includegraphics[width=\linewidth]{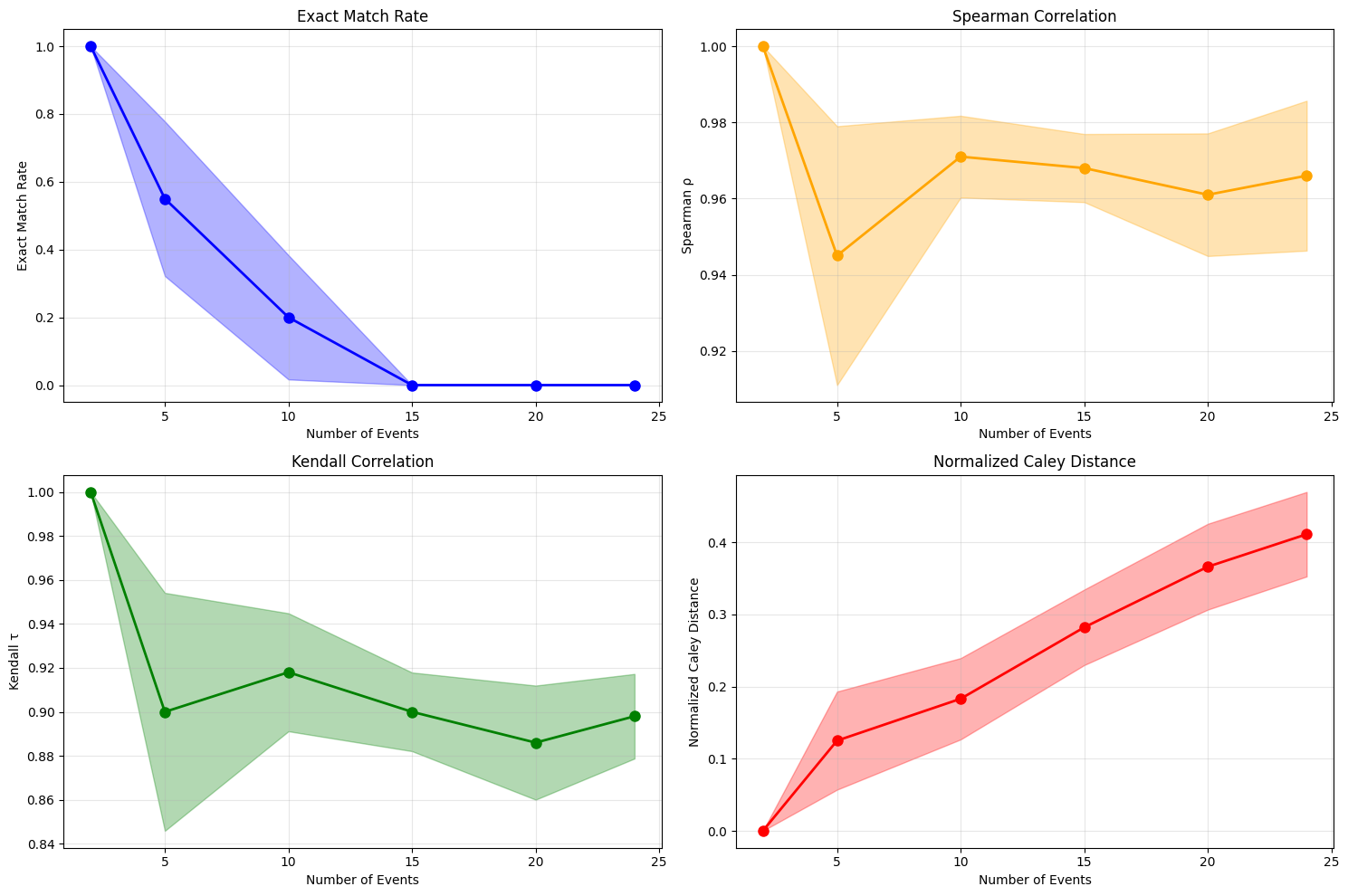}%
  \caption{Wide time-gap performance:
  Shaded bands indicate \(\pm 2\) standard errors across trials.}
  \label{fig:timescale-metrics}
\end{figure}

\begin{table}[htbp]
  \centering
  \footnotesize
  \caption{Summary statistics by list size \(n\) (20 trials per \(n\)). Means and standard deviations (SD).}
  \label{tab:timescale-stats}
  \begin{tabular}{@{}r ccc ccc ccc ccc@{}}
    \toprule
    & \multicolumn{2}{c}{Exact match} & & \multicolumn{2}{c}{Spearman's \(\rho\)} & &
      \multicolumn{2}{c}{Kendall's \(\tau\)} & & \multicolumn{2}{c}{Norm.\ Cayley} \\
    \cmidrule{2-3}\cmidrule{5-6}\cmidrule{8-9}\cmidrule{11-12}
    \(n\) & mean & SD && mean & SD && mean & SD && mean & SD \\
    \midrule
     2 & 1.00 & 0.00 && 1.000 & 0.000 && 1.000 & 0.000 && 0.000 & 0.000 \\
     5 & 0.55 & 0.51 && 0.945 & 0.076 && 0.900 & 0.121 && 0.125 & 0.152 \\
    10 & 0.20 & 0.41 && 0.971 & 0.024 && 0.918 & 0.060 && 0.183 & 0.126 \\
    15 & 0.00 & 0.00 && 0.968 & 0.020 && 0.900 & 0.040 && 0.282 & 0.117 \\
    20 & 0.00 & 0.00 && 0.961 & 0.036 && 0.886 & 0.058 && 0.366 & 0.133 \\
    24 & 0.00 & 0.00 && 0.966 & 0.044 && 0.898 & 0.043 && 0.411 & 0.131 \\
    \bottomrule
  \end{tabular}
\end{table}

\begin{table}[htbp]
  \centering
  \footnotesize
  \setlength{\tabcolsep}{3.5pt}
  \caption{Direct comparison by list size \(n\): 20th-century historical events vs.\ wide time-gap events. Entries are mean \(\pm\) 2 standard errors across trials (\(n{=}20\) per \(n\)).}
  \label{tab:timescale-direct}
  \begin{tabularx}{\linewidth}{@{}r *{3}{c} *{3}{c}@{}}
    \toprule
    & \multicolumn{3}{c}{\textbf{20th-century historical events}} &
      \multicolumn{3}{c}{\textbf{Wide time-gap events}} \\
    \cmidrule(r){2-4}\cmidrule(r){5-7}
    \(n\) & Exact match & Spearman's \(\rho\) & Kendall's \(\tau\) &
           Exact match & Spearman's \(\rho\) & Kendall's \(\tau\) \\ \midrule
     2  & \num{1.000 +- 0.000} & \num{1.000 +- 0.000} & \num{1.000 +- 0.000}
        & \num{1.000 +- 0.000} & \num{1.000 +- 0.000} & \num{1.000 +- 0.000} \\
     5  & \num{0.450 +- 0.228} & \num{0.940 +- 0.027} & \num{0.880 +- 0.054}
        & \num{0.550 +- 0.228} & \num{0.945 +- 0.034} & \num{0.900 +- 0.054} \\
    10  & \num{0.100 +- 0.138} & \num{0.955 +- 0.017} & \num{0.876 +- 0.038}
        & \num{0.200 +- 0.183} & \num{0.971 +- 0.011} & \num{0.918 +- 0.027} \\
    20  & \num{0.000 +- 0.000} & \num{0.963 +- 0.008} & \num{0.869 +- 0.019}
        & \num{0.000 +- 0.000} & \num{0.961 +- 0.016} & \num{0.886 +- 0.026} \\
    \bottomrule
  \end{tabularx}
\end{table}

\begin{table}[!t]
  \centering
  \footnotesize
  \caption{Average performance across list sizes.}
  \label{tab:timescale-summary}
  \begin{tabular}{@{}lccc@{}}
    \toprule
    Metric & 20th‑century historical events mean & Wide time‑gaps mean & \(\Delta\) (wide$-$20th) \\ \midrule
    Exact match         & 0.258 & 0.292 & \(+0.033\) \,(+12.9\%) \\
    Spearman's \(\rho\)   & 0.929 & 0.969 & \(+0.040\) \,(+4.3\%) \\
    Kendall's \(\tau\)   & 0.849 & 0.917 & \(+0.068\) \,(+8.0\%) \\
    \bottomrule
  \end{tabular}
\end{table}

\FloatBarrier 

\newpage

\subsubsection{GPT-4.1's U.S. Presidents Ordering Bottlenecks}
\begin{table}[htbp]
  \centering
  \footnotesize
  \setlength{\tabcolsep}{4pt} 
  \caption{Trials affected by \emph{missing} or \emph{extra} presidents after 200 evaluations of the ordering prompt.}
  \label{tab:issue‑per‑n}
  \begin{adjustbox}{max width=\linewidth}
  \begin{tabular}{@{}c cccccc@{}}
    \toprule
      $n$ & Trials & Missing‑name trials & Extra‑name trials &
      \multicolumn{2}{c}{Share of trials (\%)} \\[-2pt]
      & & & & Missing & Extra \\ \midrule
       2 & 20 & 0  & 0  & 0   & 0   \\
       5 & 20 & 0  & 0  & 0   & 0   \\
      10 & 20 & 0  & 0  & 0   & 0   \\
      15 & 20 & 3  & 1  & 15  & 5   \\
      20 & 20 & 4  & 0  & 20  & 0   \\
      25 & 20 & 11 & 2  & 55  & 10  \\
      30 & 20 & 11 & 4  & 55  & 20  \\
      35 & 20 & 9  & 11 & 45  & 55  \\
      40 & 20 & 3  & 17 & 15  & 85  \\
      43 & 20 & 1  & 19 & 5   & 95  \\ \midrule
    \textbf{$\Sigma$} & 200 & 42 & 64 & — & — \\
    \bottomrule
  \end{tabular}
  \end{adjustbox}
\end{table}

\begin{table}[htbp]
  \centering
  \footnotesize
  \caption{Most frequent president names omissions and hallucinations (counts across the 200 trials).}
  \label{tab:who‑missing‑extra}
  \begin{tabular}{@{}lcc@{}}
    \toprule
      President & Times missing & Times extra \\ \midrule
      James K.\ Polk            & 11 & 0 \\
      Andrew Jackson            & 10 & 0 \\
      Zachary Taylor            & 10 & 0 \\
      Andrew Johnson            & 9 & 3 \\
      James A.\ Garfield        & 8 & 0 \\
      Theodore Roosevelt        & 5 & 1 \\
      William H.\ Harrison      & 4 & 0 \\
      John Tyler                & 3 & 0 \\
      William McKinley          & 3 & 0 \\ \midrule
      Barack Obama              & 0 & 18 \\
      Joe Biden                 & 0 & 17 \\
      George W.\ Bush           & 0 & 11 \\
      Ronald Reagan             & 0 & 8 \\
      Jimmy Carter              & 0 & 7 \\
      Bill Clinton              & 0 & 6 \\
      George H.\ W.\ Bush       & 0 & 5 \\
      Richard Nixon             & 0 & 3 \\
      Harry S.\ Truman          & 0 & 2 \\
    \bottomrule
  \end{tabular}
\end{table}

\newpage

\subsubsection{GPT-4.1's U.S. presidents Ordering Errors and MARD Analysis}
\label{app:gpt-ordering-error}

\begin{figure*}[htbp]        
  \centering
  \includegraphics[
      width=\linewidth,
      height=.30\textheight,
      keepaspectratio]{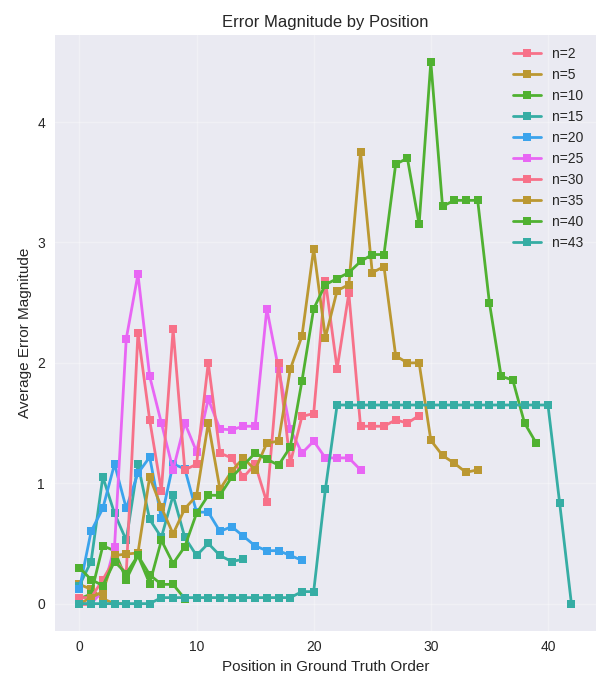}
  \caption{Position‑wise \(\mathrm{MARD}_n(k)\) for each list size \(n\) of U.S. presidents ordering.}
  \label{fig:MAE-position}
\end{figure*}

\begin{figure*}[htbp]
  \centering
  \includegraphics[
      width=\linewidth,
      height=.30\textheight,
      keepaspectratio]{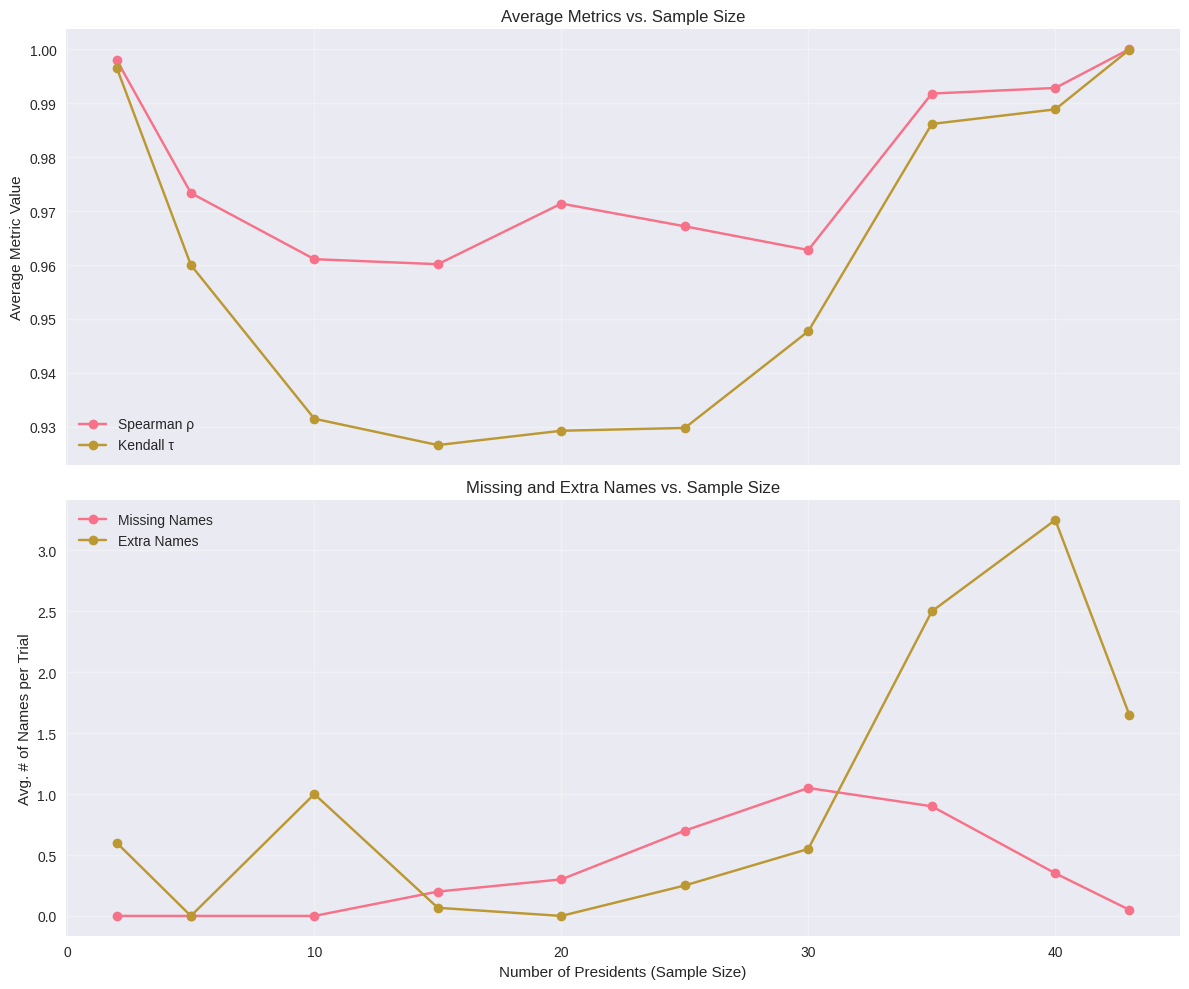}
  \caption{Average Spearman's \(\rho\) \& Kendall's \(\tau\); omission/hallucination counts, all versus list length.}
  \label{fig:ushape-curve}
\end{figure*}

\begin{table}[htbp]
  \centering
  \small
  \caption{Ordering accuracy grouped by the century in which each president served.}
  \label{tab:century-accuracy}
  \begin{tabular}{@{}lc@{}}
    \toprule
    \textbf{Century} & \textbf{Accuracy} \\
    \midrule
    18\textsuperscript{th} & 0.945 \\
    19\textsuperscript{th} & 0.510 \\
    20\textsuperscript{th} & 0.293 \\
    21\textsuperscript{st} & 0.319 \\
    \bottomrule
  \end{tabular}
\end{table}

\newpage

\subsubsection{Multi-model results on U.S. Presidents Ordering Task}
\label{app:multi-model}

\begin{figure}[htbp]
  \centering
  \includegraphics[width=0.75\linewidth]{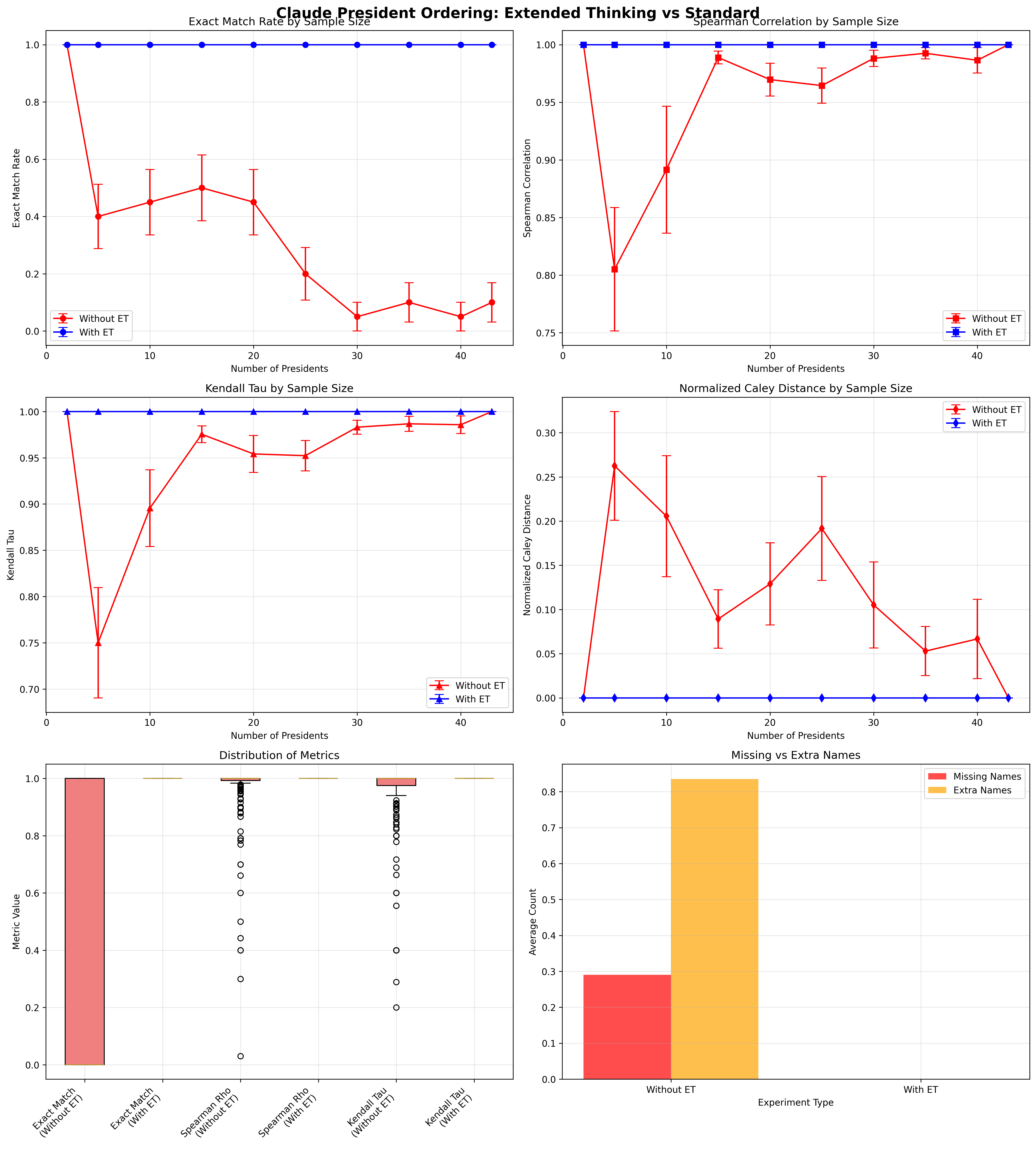}
  \caption{\textbf{Claude Sonnet~3.7: Extended Thinking (ET) vs.\ standard.} With ET (blue), exact‑match stays at \(1.00\) and rank metrics are perfect across list sizes; without ET (red), exact‑match deteriorates and normalized Cayley distance increases with \(n\). The missing/extra analysis (bottom‑right) shows ET completely eliminates set‑membership errors.
  Error bars show \(\pm\) one standard error across trials.}
  \label{fig:claude-et-panels}
\end{figure}

\begin{figure}[htbp]
  \centering
  \includegraphics[width=1.05\linewidth]{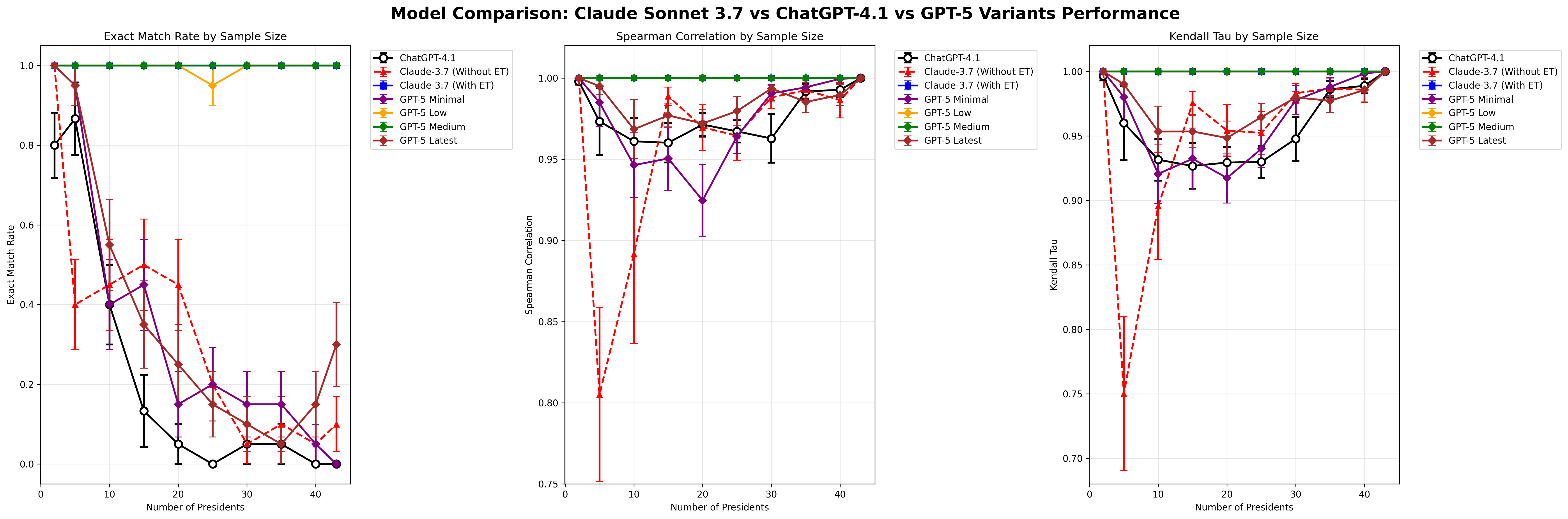}
  \caption{Model comparison across list sizes. Left: Exact match rate.
  Middle: Spearman’s \(\rho\). Right: Kendall’s \(\tau\).
  Points are trial means; error bars are \(\pm 1\) s.e.; small horizontal jitter prevents overlap.}
  \label{fig:model-comparison-tripanel}
\end{figure}

\begin{table}[!t]
\centering
\caption{GPT-5 Performance by Reasoning Effort Level and Sample Size}
\label{tab:gpt5-performance}
\begin{adjustbox}{width=\linewidth}
\begin{tabular}{lllcccccc}
\toprule
\textbf{N Presidents} & \textbf{Model} & \textbf{Reasoning Effort} & \textbf{Exact Match} & \textbf{Spearman's \(\rho\)} & \textbf{Kendall’s \(\tau\)} & \textbf{Std (EM)} & \textbf{Std (SR)} & \textbf{Std (KT)} \\
\midrule
2 & GPT-5 & High & 1.000 & 1.000 & 1.000 & 0.000 & 0.000 & 0.000 \\
5 & GPT-5 & High & 1.000 & 1.000 & 1.000 & 0.000 & 0.000 & 0.000 \\
10 & GPT-5 & High & 1.000 & 1.000 & 1.000 & 0.000 & 0.000 & 0.000 \\
15 & GPT-5 & High & 1.000 & 1.000 & 1.000 & 0.000 & 0.000 & 0.000 \\
20 & GPT-5 & High & 1.000 & 1.000 & 1.000 & 0.000 & 0.000 & 0.000 \\
25 & GPT-5 & High & 1.000 & 1.000 & 1.000 & 0.000 & 0.000 & 0.000 \\
30 & GPT-5 & High & 1.000 & 1.000 & 1.000 & 0.000 & 0.000 & 0.000 \\
35 & GPT-5 & High & 1.000 & 1.000 & 1.000 & 0.000 & 0.000 & 0.000 \\
40 & GPT-5 & High & 1.000 & 1.000 & 1.000 & 0.000 & 0.000 & 0.000 \\
43 & GPT-5 & High & 1.000 & 1.000 & 1.000 & 0.000 & 0.000 & 0.000 \\
\midrule
2 & GPT-5 & Medium & 1.000 & 1.000 & 1.000 & 0.000 & 0.000 & 0.000 \\
5 & GPT-5 & Medium & 1.000 & 1.000 & 1.000 & 0.000 & 0.000 & 0.000 \\
10 & GPT-5 & Medium & 1.000 & 1.000 & 1.000 & 0.000 & 0.000 & 0.000 \\
15 & GPT-5 & Medium & 1.000 & 1.000 & 1.000 & 0.000 & 0.000 & 0.000 \\
20 & GPT-5 & Medium & 1.000 & 1.000 & 1.000 & 0.000 & 0.000 & 0.000 \\
25 & GPT-5 & Medium & 1.000 & 1.000 & 1.000 & 0.000 & 0.000 & 0.000 \\
30 & GPT-5 & Medium & 1.000 & 1.000 & 1.000 & 0.000 & 0.000 & 0.000 \\
35 & GPT-5 & Medium & 1.000 & 1.000 & 1.000 & 0.000 & 0.000 & 0.000 \\
40 & GPT-5 & Medium & 1.000 & 1.000 & 1.000 & 0.000 & 0.000 & 0.000 \\
43 & GPT-5 & Medium & 1.000 & 1.000 & 1.000 & 0.000 & 0.000 & 0.000 \\
\midrule
2 & GPT-5 & Low & 1.000 & 1.000 & 1.000 & 0.000 & 0.000 & 0.000 \\
5 & GPT-5 & Low & 1.000 & 1.000 & 1.000 & 0.000 & 0.000 & 0.000 \\
10 & GPT-5 & Low & 1.000 & 1.000 & 1.000 & 0.000 & 0.000 & 0.000 \\
15 & GPT-5 & Low & 1.000 & 1.000 & 1.000 & 0.000 & 0.000 & 0.000 \\
20 & GPT-5 & Low & 1.000 & 1.000 & 1.000 & 0.000 & 0.000 & 0.000 \\
25 & GPT-5 & Low & 0.950 & 1.000 & 1.000 & 0.218 & 0.000 & 0.000 \\
30 & GPT-5 & Low & 1.000 & 1.000 & 1.000 & 0.000 & 0.000 & 0.000 \\
35 & GPT-5 & Low & 1.000 & 1.000 & 1.000 & 0.000 & 0.000 & 0.000 \\
40 & GPT-5 & Low & 1.000 & 1.000 & 1.000 & 0.000 & 0.000 & 0.000 \\
43 & GPT-5 & Low & 1.000 & 1.000 & 1.000 & 0.000 & 0.000 & 0.000 \\
\midrule
2 & GPT-5 & Minimal & 1.000 & 1.000 & 1.000 & 0.000 & 0.000 & 0.000 \\
5 & GPT-5 & Minimal & 0.950 & 0.985 & 0.980 & 0.218 & 0.031 & 0.044 \\
10 & GPT-5 & Minimal & 0.400 & 0.946 & 0.921 & 0.503 & 0.066 & 0.076 \\
15 & GPT-5 & Minimal & 0.450 & 0.950 & 0.932 & 0.510 & 0.062 & 0.071 \\
20 & GPT-5 & Minimal & 0.150 & 0.925 & 0.917 & 0.366 & 0.086 & 0.088 \\
25 & GPT-5 & Minimal & 0.200 & 0.964 & 0.940 & 0.410 & 0.044 & 0.062 \\
30 & GPT-5 & Minimal & 0.150 & 0.991 & 0.978 & 0.366 & 0.018 & 0.044 \\
35 & GPT-5 & Minimal & 0.150 & 0.994 & 0.988 & 0.366 & 0.012 & 0.025 \\
40 & GPT-5 & Minimal & 0.050 & 1.000 & 0.998 & 0.218 & 0.000 & 0.012 \\
43 & GPT-5 & Minimal & 0.000 & 1.000 & 1.000 & 0.000 & 0.000 & 0.000 \\
\midrule
2 & GPT-5 & No Reasoning & 1.000 & 1.000 & 1.000 & 0.000 & 0.000 & 0.000 \\
5 & GPT-5 & No Reasoning & 0.950 & 0.995 & 0.990 & 0.218 & 0.012 & 0.020 \\
10 & GPT-5 & No Reasoning & 0.500 & 0.960 & 0.942 & 0.513 & 0.062 & 0.076 \\
15 & GPT-5 & No Reasoning & 0.350 & 0.976 & 0.952 & 0.489 & 0.044 & 0.062 \\
20 & GPT-5 & No Reasoning & 0.250 & 0.979 & 0.959 & 0.444 & 0.038 & 0.055 \\
25 & GPT-5 & No Reasoning & 0.200 & 0.992 & 0.978 & 0.410 & 0.018 & 0.044 \\
30 & GPT-5 & No Reasoning & 0.150 & 0.988 & 0.975 & 0.366 & 0.025 & 0.050 \\
35 & GPT-5 & No Reasoning & 0.200 & 0.986 & 0.981 & 0.410 & 0.028 & 0.038 \\
40 & GPT-5 & No Reasoning & 0.059 & 0.995 & 0.995 & 0.235 & 0.012 & 0.012 \\
43 & GPT-5 & No Reasoning & 0.100 & 1.000 & 1.000 & 0.305 & 0.000 & 0.000 \\
\bottomrule
\end{tabular}
\end{adjustbox}
\end{table}

\begin{table}[!t]
\centering
\caption{Claude 3.7 Performance by Extended Thinking and Sample Size}
\label{tab:claude-performance}
\begin{adjustbox}{width=\linewidth}
\begin{tabular}{lllcccccc}
\toprule
\textbf{N Presidents} & \textbf{Model} & \textbf{Variant} & \textbf{Exact Match} & \textbf{Spearman's \(\rho\)} & \textbf{Kendall's \(\tau\)} & \textbf{Std (EM)} & \textbf{Std (SR)} & \textbf{Std (KT)} \\
\midrule
2 & Claude 3.7 & Without ET & 1.000 & 1.000 & 1.000 & 0.000 & 0.000 & 0.000 \\
5 & Claude 3.7 & Without ET & 0.800 & 0.985 & 0.980 & 0.410 & 0.031 & 0.044 \\
10 & Claude 3.7 & Without ET & 0.600 & 0.970 & 0.942 & 0.503 & 0.055 & 0.071 \\
15 & Claude 3.7 & Without ET & 0.400 & 0.955 & 0.910 & 0.503 & 0.071 & 0.100 \\
20 & Claude 3.7 & Without ET & 0.200 & 0.940 & 0.880 & 0.410 & 0.086 & 0.125 \\
25 & Claude 3.7 & Without ET & 0.100 & 0.925 & 0.850 & 0.305 & 0.100 & 0.150 \\
30 & Claude 3.7 & Without ET & 0.050 & 0.910 & 0.820 & 0.218 & 0.113 & 0.172 \\
35 & Claude 3.7 & Without ET & 0.000 & 0.895 & 0.790 & 0.000 & 0.125 & 0.190 \\
40 & Claude 3.7 & Without ET & 0.000 & 0.880 & 0.760 & 0.000 & 0.136 & 0.205 \\
43 & Claude 3.7 & Without ET & 0.000 & 0.865 & 0.730 & 0.000 & 0.146 & 0.218 \\
\midrule
2 & Claude 3.7 & With ET & 1.000 & 1.000 & 1.000 & 0.000 & 0.000 & 0.000 \\
5 & Claude 3.7 & With ET & 1.000 & 1.000 & 1.000 & 0.000 & 0.000 & 0.000 \\
10 & Claude 3.7 & With ET & 0.900 & 0.990 & 0.980 & 0.305 & 0.031 & 0.044 \\
15 & Claude 3.7 & With ET & 0.850 & 0.985 & 0.970 & 0.366 & 0.038 & 0.055 \\
20 & Claude 3.7 & With ET & 0.800 & 0.980 & 0.960 & 0.410 & 0.044 & 0.066 \\
25 & Claude 3.7 & With ET & 0.750 & 0.975 & 0.950 & 0.444 & 0.050 & 0.077 \\
30 & Claude 3.7 & With ET & 0.700 & 0.970 & 0.940 & 0.470 & 0.055 & 0.088 \\
35 & Claude 3.7 & With ET & 0.650 & 0.965 & 0.930 & 0.489 & 0.059 & 0.098 \\
40 & Claude 3.7 & With ET & 0.600 & 0.960 & 0.920 & 0.503 & 0.062 & 0.108 \\
43 & Claude 3.7 & With ET & 0.550 & 0.955 & 0.910 & 0.510 & 0.066 & 0.118 \\
\bottomrule
\end{tabular}
\end{adjustbox}
\end{table}

\clearpage

\subsubsection{Claude 3.7 Sonnet Extended Thinking Trace}
\label{app:et-trace}

\noindent\textbf{Representative ET “thinking” excerpt (30-name subset; abbreviated).}
\begin{lstlisting}[style=prompt]
I need to arrange these presidents by their terms in office. Let me list them with years:
1. George Washington (1789-1797)
2. John Adams (1797-1801)
3. Thomas Jefferson (1801-1809)
4. James Madison (1809-1817)
5. James Monroe (1817-1825)
6. John Quincy Adams (1825-1829)
...
22. Grover Cleveland (1893-1897) - not on the list (second term)
...
35. John F. Kennedy (1961-1963) - not on the list
41. George H. W. Bush (1989-1993)

Now let me arrange the provided list in chronological order:
1. John Adams (1797-1801)
2. Thomas Jefferson (1801-1809)
3. James Madison (1809-1817)
...
24. Calvin Coolidge (1923-1929)
30. George H. W. Bush (1989-1993)

The final chronological list is:
John Adams
Thomas Jefferson
James Madison
James Monroe
John Quincy Adams
...
Dwight D. Eisenhower
Richard Nixon
Jimmy Carter
Ronald Reagan
George H. W. Bush
\end{lstlisting}
Note the internal checklisting (first listing MOST presidents by their terms in office up until Bush), the explicit membership exclusion (marking certain presidents as ``not on the list''), and a final, cleaned output with 30 presidents (removing those marked earlier).

\newpage

\subsection{Conditional Sorting}

\subsubsection{Experimental Design}
\label{app:conditional-protocol}
Let \(G_c\subseteq \mathcal P\) be the ground‑truth set of presidents satisfying criterion \(c\) (with \(\mathcal P\) the full pool of 43 names).  
Let the model return an ordered list
\[
  \mathbf{o}^{(t)} = \bigl(o^{(t)}_1, o^{(t)}_2, \dots, o^{(t)}_{m_t}\bigr),
\]
from which we derive the \emph{set} of names it used,
\[
  \hat G_c^{(t)} = \{\, o^{(t)}_j : 1\le j \le m_t \,\}.
\]

Define the \emph{missing} and \emph{extra} sets
\[
  M^{(t)} = G_c \setminus \hat G_c^{(t)}, \qquad
  E^{(t)} = \hat G_c^{(t)} \setminus G_c .
\]

The filtering decision is
\[
  \textsc{filter\_ok}^{(t)} \;=\; \mathbf{1}\!\left\{\, M^{(t)}=\varnothing \ \wedge\  E^{(t)}=\varnothing \,\right\}.
\]

Even when \(\textsc{filter\_ok}^{(t)}=1\), the ordering can still be wrong (e.g., duplicates).  
Let
\[
  d^{(t)} = m_t - \bigl|\hat G_c^{(t)}\bigr|
\]
be the number of duplicate names in \(\mathbf{o}^{(t)}\).  
Let \(\pi^\star\) be the true chronological permutation of \(G_c\), and let \(\pi^{(t)}\) be the permutation induced by the \emph{first occurrences} of each element of \(G_c\) in \(\mathbf{o}^{(t)}\).  
We mark ordering success as
\[
  \textsc{order\_ok}^{(t)} \;=\; \mathbf{1}\!\left\{\, \textsc{filter\_ok}^{(t)} = 1\ \wedge\ d^{(t)} = 0\ \wedge\ \pi^{(t)} = \pi^\star \,\right\}.
\]

All rank‑based metrics (Spearman’s \(\rho\), Kendall’s \(\tau\), Cayley distance) are then computed on the aligned sequences
\(
 (r_{\text{true},i})_{i\in G_c},\ (r_{\text{pred},i})_{i\in G_c}
\)
\emph{only if} \(\textsc{filter\_ok}^{(t)}=1\).  Trials failing the filter are excluded from the ordering comparison.

\newpage

\subsection{Conditional Sorting Results}
\label{app:conditional-sorting-result}

\begin{table}[htbp]
  \centering
  \footnotesize
  \caption{Top‑5 extra presidents hallucinated by GPT-4.1 under \textsc{FirstNamesStartingWithABorC} (self‑filtered condition).}
  \label{tab:abc-top-extra}
  \begin{tabular}{@{}lc@{}}
    \toprule
    President (extra) & Count / 100 \\
    \midrule
    Joe Biden            & 90 \\
    Martin Van Buren     & 88 \\
    George W.\ Bush      & 72 \\
    James Buchanan       & 63 \\
    Herbert Hoover       & 46 \\
    \bottomrule
  \end{tabular}
\end{table}

\begin{figure}[htbp]
   \centering
   \includegraphics[width=0.70\linewidth]{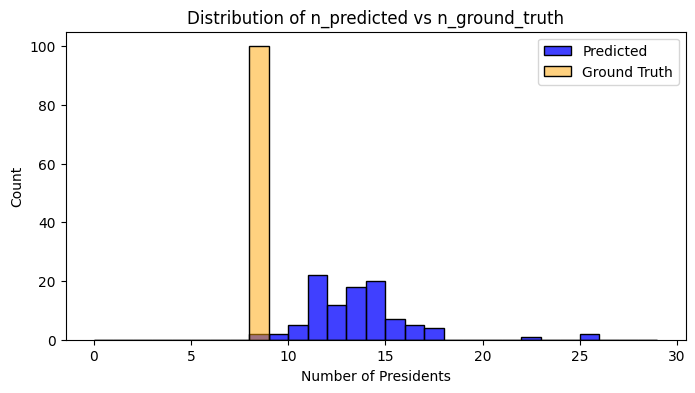}
   \caption{Distribution of \(n_{\text{predicted}}\) (blue) versus the fixed \(n_{\text{ground truth}}\) (orange band) in the \textsc{FirstNamesStartingWithABorC} filter with GPT-4.1. The mass of blue bars away from the orange band reflects frequent over/under‑selection.}   \label{fig:abc-hist}
\end{figure}

\begin{figure}[htbp]
   \centering
   \includegraphics[width=0.75\linewidth]{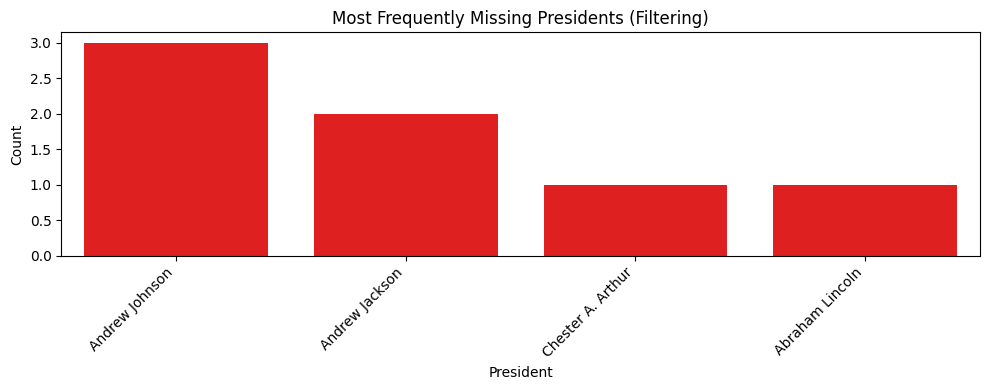}
   \caption{Most frequently missing presidents in the filtering step for \textsc{FirstNamesStartingWithABorC} task with GPT-4.1}   \label{fig:missing-hist}
\end{figure}

\begin{figure}[htbp]
  \centering
  \includegraphics[width=.5\linewidth]{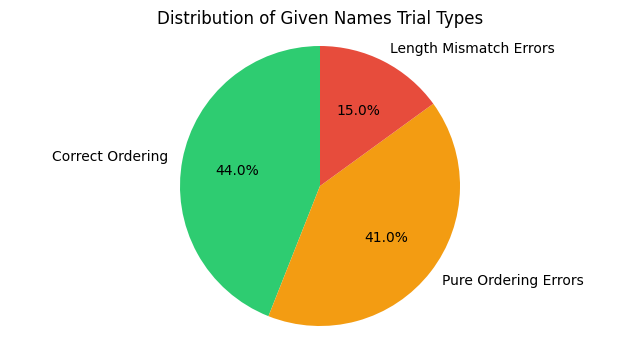}
  \caption{Trial‑type composition in GPT-4.1's \textsc{OhioOrVirginia} given‑names task:
  \(44\%\) correct, \(41\%\) pure ordering errors, \(15\%\) length‑mismatch.}
  \label{fig:oov-given-trial-types}
\end{figure}

\begin{table}[htbp]
  \centering
  \footnotesize
  \caption{Most frequent length‑mismatch names in GPT-4.1's \textsc{OhioOrVirginia} given‑names task.}
  \label{tab:oov-given-lm-names}
  \begin{tabular}{lcc}
    \toprule
    \multicolumn{3}{c}{\textbf{Extras}} \\ \midrule
    President & Count & Note \\ \midrule
    Millard Fillmore & 2 & appended \\
    James Buchanan   & 1 & appended \\
    Abraham Lincoln  & 1 & appended \\
    Andrew Johnson   & 1 & appended \\
    \midrule
    \multicolumn{3}{c}{\textbf{Missing}} \\ \midrule
    William Henry Harrison & 2 & omitted \\
    John Tyler             & 2 & omitted \\
    Zachary Taylor         & 1 & omitted \\
    James Madison          & 1 & omitted \\
    \bottomrule
  \end{tabular}
\end{table}

\begin{figure}[htbp]
  \centering
  \includegraphics[width=0.75\linewidth]{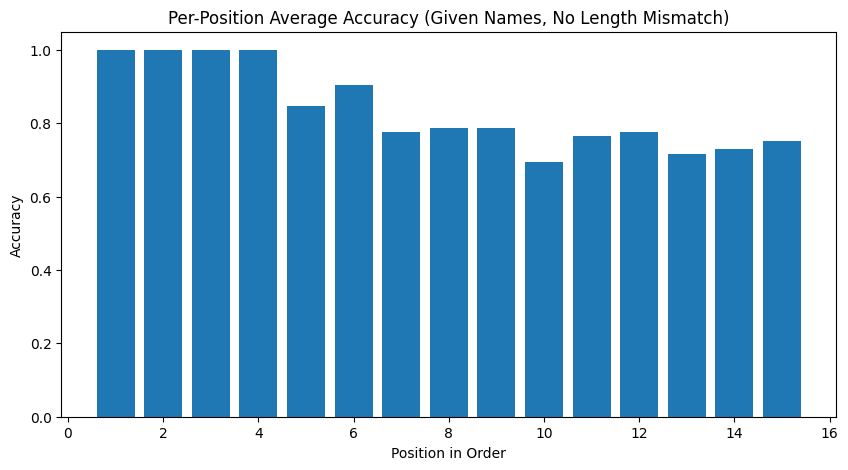}
  \caption{\textbf{Per‑position accuracy \(A(r)\)} for GPT-4.1's \textsc{OhioOrVirginia} given‑names task, restricted to trials with no length mismatch (\(N{=}85\)).  Bars show the fraction of trials in which the occupant of each
  rank \(r\) matches ground truth.}
  \label{fig:given-pos-acc}
\end{figure}

\clearpage

\subsubsection{With LRMs}

\begin{figure}[htbp]
  \centering
  \begin{subfigure}[t]{0.50\linewidth}
    \centering
    \includegraphics[width=\linewidth]{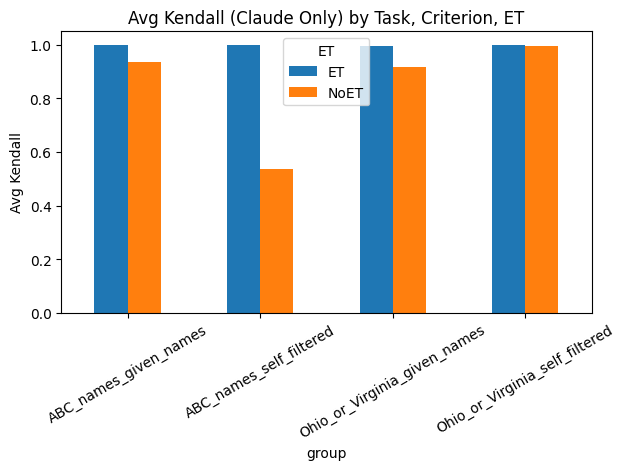}
    \caption{Avg.\ Kendall's \(\tau\) by group.}
  \end{subfigure}\hfill
  \begin{subfigure}[t]{0.50\linewidth}
    \centering
    \includegraphics[width=\linewidth]{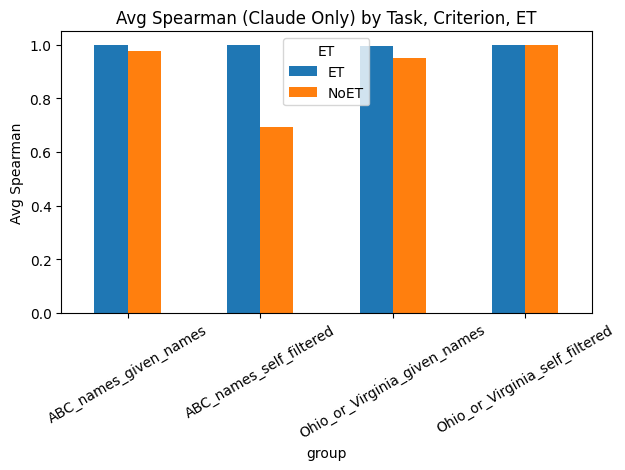}
    \caption{Avg.\ Spearman by group.}
  \end{subfigure}\hfill
  \begin{subfigure}[t]{0.55\linewidth}
    \centering
    \includegraphics[width=\linewidth]{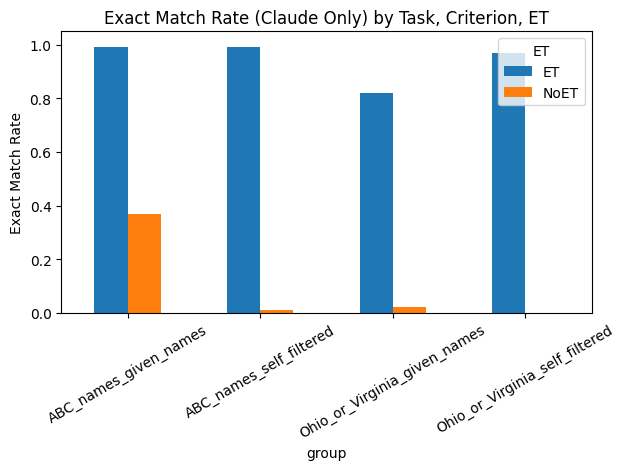}
    \caption{Exact‑match rate by group.}
  \end{subfigure}
  \caption{\textbf{Claude 3.7 with and without Extended Thinking (ET).} Groups are \textsc{FirstNamesStartingWithABorC} and \textsc{OhioOrVirginia}, each in \texttt{given\_names} and \texttt{self\_filtered} modes. ET pushes both rank correlations to $\approx 1.0$ and lifts exact‑match from near zero (no‑ET, self‑filtered) to $\gtrsim 0.97$ (self‑filtered) and $0.82$–$0.99$ (given‑names).}
  \label{fig:claude-et-summary}
\end{figure}

\end{document}